 \documentclass[final,5p,times,twocolumn]{elsarticle}

\usepackage{enumitem}
\usepackage{url}
\usepackage[utf8]{inputenc}
\usepackage{xcolor}
\usepackage{natbib}

\usepackage{array} 
\usepackage{amsmath}
\renewcommand{\arraystretch}{1.5}

\usepackage{amssymb}

\usepackage{hyperref}
\usepackage{wrapfig}
\usepackage{subfig}
\usepackage{appendix}
\usepackage{caption}
\usepackage{booktabs}
\usepackage{array}
\usepackage{makecell}
\usepackage{multicol}
\usepackage{caption}
\usepackage{geometry}
\usepackage{multirow}
\usepackage{graphicx}
\usepackage{float}
\usepackage{subcaption}
\usepackage{comment}
\usepackage{tabularx} 

\journal{.}

\begin{document}

\begin{frontmatter}



\title{Quantile deep learning models for multi-step ahead time series prediction}
 
\author[inst1]{Jimmy Cheung\corref{contrib}}

\author[inst1]{Smruthi Rangarajan\corref{contrib}}

\author[inst1]{Amelia Maddocks\corref{contrib}} 

\author[inst1]{Xizhe Chen
 }

\author[inst1]{Rohtiash Chandra\corref{corauthor}}
\ead{rohitash.chandra@unsw.edu.au}

\cortext[corauthor]{Corresponding author}
\cortext[contrib]{Joint first authors with equal contribution.}

\affiliation[inst1]{organization={Transitional Artificial Intelligence Research Group, School of Mathematics and Statistics}, 
            addressline={ University of New South Wales}, 
            city={Sydney},  
            country={Australia}}
            
\begin{abstract}
    Uncertainty quantification is crucial in time series prediction, and quantile regression offers a valuable mechanism for uncertainty quantification which is useful for extreme value forecasting. Although deep learning models have been prominent in multi-step ahead prediction, the development and evaluation of quantile deep learning models have been limited.  We present a novel quantile regression deep learning framework for multi-step time series prediction. In 
 this way, we elevate the capabilities of deep learning models by incorporating quantile regression, thus providing a more nuanced understanding of predictive values. We provide an implementation of prominent deep learning models for multi-step ahead time series prediction and evaluate their performance under high volatility and extreme conditions. 
     We include multivariate and univariate modelling, strategies and provide a comparison with conventional deep learning models from the literature. 
    Our models are tested on two cryptocurrencies: Bitcoin and Ethereum, using daily close-price data and selected benchmark time series datasets. The results show that integrating a quantile loss function with deep learning provides additional predictions for selected quantiles without a loss in the prediction accuracy when compared to the literature. Our quantile model has the ability to handle volatility more effectively and provides additional information for decision-making and uncertainty quantification through the use of quantiles when compared to conventional deep learning models.



\end{abstract}


\end{frontmatter}


\section{Introduction}
\label{sec:sample1}

In the realm of time series forecasting, uncertainty quantification is a critical component that allows for more informed decision-making, particularly in fields characterised by high volatility such as financial markets, energy demand, and weather forecasting. Conventional deep learning models, whilst powerful in multi-step ahead forecasting, often fall short in providing comprehensive measures of uncertainty. This gap can be addressed by integrating quantile regression, a statistical technique that offers a mechanism for extreme forecasting by predicting the conditional quantiles of a response variable. Koenker and Bassett \cite{Koenker1978econometrica} introduced the quantile regression model in the mid-1970s to estimate conditional quantiles, offering a measure of uncertainty rather than single-point predictions as in conventional linear regression models. Quantile regression has been widely used in statistical analysis \cite{Waldmann2018quantile} and finds applications in various fields, including epidemiology \cite{wei2019applications}, economics \cite{Fitzenberger2013economic,Tillaguango2024impact}, ecology \cite{briollais2014application}, and finance \cite{allen2009quantile}. For instance, in the field of economics, it has been employed to study salary distributions influenced by returns to education and student experience \cite{buchinsky1994changes}. In medicine, quantile regression has been used to analyse the effects of different local anesthetics on the duration of nerve blocks \cite{staffa2019quantile}. Unlike traditional linear models such as least squares regression \cite{watson1967linear,geladi1986partial}, quantile regression provides more comprehensive information about the conditional distribution, revealing data characteristics across different quantiles as well as the average of the data. Hence, this approach offers a methodology for projecting uncertainties in prediction \cite{Vaysse2017quantile, Dogulu2015estimation}.

Extreme value prediction \cite{ochi1973prediction} focuses on forecasting rare and significant events, these are often outliers or extreme values in a dataset and have a low probability of occurrence but can cause major consequences \cite{ribeiro2020imbalanced}. In a meteorology context, an example is the rapid intensification of cyclones \cite{wang2008climate}. Given the distribution of a dataset, this statistical modeling approach targets the tail of the distribution where extreme events reside, allowing for the estimation of their probability. It finds applications in various fields, including natural disasters \cite{makkonen2008problems, lee2021temporal, xing2020casualty}, financial crises \cite{zhao2014extreme}, and system failures \cite{wang2024quantifying}. Extreme value prediction is crucial for enhancing risk management as it provides a foundation for developing emergency plans \cite{cumperayot2013early} and preventive measures \cite{messervey2019application}. For instance, it can be used to assess potential casualties in earthquake disasters \cite{xing2020casualty}, helping to minimise losses. Additionally, whilst extreme value prediction targets the rare, extreme events that in the tails of the distribution, quantile regression provides a more generalised approach to estimate various quantiles, such as the median and 90$^{\text{th}}$ percentile of the response variable's conditional distribution \cite{cai2013extreme}. However, classic quantile regression can perform poorly for extreme values. When integrated with extreme value theory (EVT), extreme quantile regression can estimate conditional quantiles that extend beyond the observed data range \cite{velthoen2023gradient}. Further literature on extreme value prediction using a quantile function model are detailed by Cai \textit{et al.} \cite{cai2013extreme}.


Deep learning models can handle complex, high-dimensional data and extract hidden patterns and features, making them particularly effective for time series prediction, especially with nonlinear and multivariate data \cite{lim2021time}. These models have been extensively used for time series forecasting, including univariate, multivariate, single-step, and multi-step predictions \cite{torres2021deep}. In the meteorological field, deep learning can be used to predict extreme weather events \cite{fang2021survey} such as smog \cite{chen2020evanet}, heavy rainfall \cite{gope2016early}, and declining groundwater levels \cite{chen2024deep}. By analysing historical meteorological data and satellite images, deep learning models can also identify early signals of extreme weather, enabling advanced preparation to mitigate potential damage \cite{fang2021survey}.

The combination of quantile regression with deep learning is gaining traction \cite{Papacharalampous2024uncertainty, Tyralis2024a}. Deep learning-based quantile regression has been applied to right-censored survival data \cite{Jia2022deep}, utilising the Huber check function and inverse probability censoring weights (IPCW) function to more accurately adjust for censoring. This has been validated through simulation studies and applications to breast cancer gene datasets \cite{Jia2022deep}.

Recent studies have utilised the quantile regression forests (QRF) model to predict road traffic volume, showing significant implications for regional development \cite{Van2024a}. Furthermore, integrating quantile regression with deep learning models, such as the long short-term memory (LSTM) network has significantly improved the accuracy and reliability of river runoff predictions \cite{zhu2024probabilistic}. The monotone quantile regression neural network (MQRNN) was employed by Hu \textit{et al.} \cite{Hu2024a} to address the quantile crossing problem in time series prediction by taking the monotonicity of quantile into consideration. An improved quantile regression neural network (iQRNN) \cite{Zhang2018improved} was used for probabilistic load forecasting that utilised deep learning strategies such as batch training, early stopping, and dropout regularisation that significantly improved the training efficiency and prediction stability of the model. Recent advancements including MQRNN \cite{Hu2023novel} and the deep partially linear quantile regression neural network (DPLQR) model \cite{Tang2023neural} have addressed  quantile crossover, where different quantile estimation lines (e.g. 10\% quantile, 50\% quantile) may cross or stagger during the prediction process in the quantile regression, leading to inconsistency or irrationality in the prediction results, and constructing confidence intervals for time series predictions. These models highlight the potential of combining deep learning with quantile regression for enhanced uncertainty quantification.

The integration of deep learning and extreme value prediction has demonstrated significant application potential across various fields. Deep learning has been leveraged to predict extreme market fluctuations, aiding investors anticipate the risk of financial crises or market crashes \cite{Melina2023a}. The combination of extreme value theory (EVT) with neural networks has significantly improved the accuracy of predicting extreme events in financial markets \cite{Ibn2018the}. A hybrid model framework that combines EVT and machine learning \cite{Melina2024modeling} can more accurately estimate stock market risks by processing multivariate and high-frequency data, thereby enhancing risk management and investment decision-making accuracy. Furthermore, there is limited work in the area of quantile regression for multi-step ahead forecasting. 

In this study, we present a novel quantile regression deep learning framework for multi-step time series prediction. In 
 this way, we elevate the capabilities of deep learning models by incorporating quantile regression, thus providing a more nuanced understanding of predictive values.  We evaluate the framework using  univariate and multivariate benchmark datasets and focus on multi-step ahead time series predictions under conditions of high volatility and extremes that include cryptocurrency market, specifically Bitcoin and Ethereum datasets used by Wu \textit{et al.} \cite{Wu2024review}. We evaluate the framework with two novel deep learning models that include LSTM networks \cite{hochreiter1997long} and convolutional neural networks \cite{alzubaidi2021review} which have been very promising for multi-step ahead forecasting \cite{chandra2021evaluation}. We provide open-source Python code and data so that our framework can be extended and applied to various fields that feature extreme values and require uncertainty quantification. 

The rest of the paper is organised as follows. In Section 2, we provide background and related work, and in Section 3, we present the methodology. Section 4 presents the results and Section 5 and 6 provide the discussion and conclusions, respectively. 



\section{Background and Related Work}

\subsection{Quantile regression}
The quantile regression model is an extension of linear regression  that estimates the conditional median (other quantiles) of the response variable using the conditional quantile function. 
For the $\tau$-th quantile ($0 < \tau < 1$), the quantile model is:
\begin{equation}
Q_y(\tau | X) = X\beta(\tau)
\end{equation}
where:
$Q_y(\tau | X)$ represents the $\tau$-th quantile of the dependent variable $y$ given the independent variables $X$.
$X$ is the vector of independent variables, containing $n$ observations.
$\beta(\tau)$ is the vector of coefficients associated with the $\tau$-th quantile.
Quantile regression estimates the conditional distribution of the dependent variables under the different quantiles. It is different from ordinary least squares regression (OLS) which is based on the conditional mean of the estimated dependent variable. Therefore, quantile regression can provide a more comprehensive understanding of the data and provide a means of uncertainty quantification in predictions. Through quantile regression, we can obtain a more detailed description of the entire distribution of a given dataset by estimating different quantiles (such as the 10$^{\text{th}}$, 50$^{\text{th}}$, and 90$^{\text{th}}$ quantiles).

Quantile regression is less sensitive to outliers because its estimation is based on minimising the absolute error with a specialised loss function, rather than the conventional squared error loss in linear models. The quantile loss function is given as follows:

 \begin{equation}
\rho_\tau(u) = u(\tau - 1_{u < 0)})
\end{equation}

where: $u = y - X\beta(\tau)$ represents the residuals.
$\rho_\tau(u)$ is the asymmetric absolute loss function, also known as the "check loss function."  
$I(\cdot)$ is the indicator function, which takes the value 1 if the condition inside is true, otherwise 0.

\begin{figure}[htbp!]
    \centering
    \includegraphics[width=0.8\linewidth]{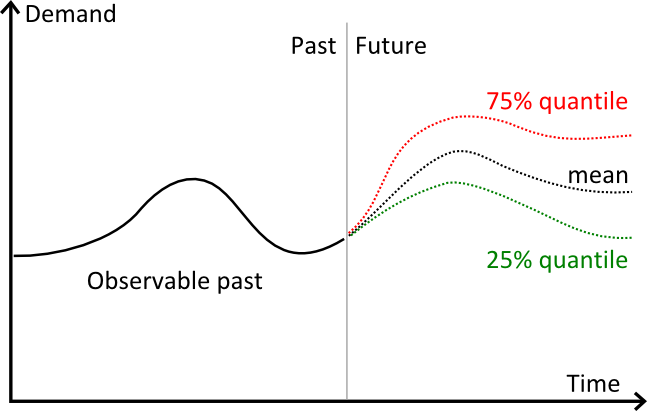}
    \caption{The uncertainty range provided in the demand forecast using quantile regression includes the forecast mean and the trend of the upper and lower quantiles (25\% and 75\%).}
    \label{fig:quantiles}
\end{figure}

 A simple linear regression example 
 can be used to visualise the concept of quantiles, where regression lines for different quantiles are displayed on the same graph as shown in Figure \ref{fig:quantiles}.

Quantile regression is also know to be applicable for data that is heteroskedastic, providing for a more precise estimate. Moreover, quantile regression can capture the potential non-linear and heterogeneous effects of explanatory variables on the dependent variable, allowing for enhanced insights on the given dataset \cite{Zhu2018heterogeneous, Koenker2004quantile}.

The application of quantile regression in time series analysis has garnered widespread attention \cite{Cai2002regression} by expanding modelling options through allowing for modelling of local and quantile-specific dynamics. For example, the weighted Nadaraya-Watson (WNW) regression is a novel method to implement quantile regression, it effectively estimates conditional quantiles in time series data \cite{Cai2002regression}. Additionally, quantile regression can be applied to interval forecasting, structural change detection, and portfolio construction \cite{Xiao2012time}. 
Koenker \cite{Koenker2017quantile} provided a comprehensive review on the development and applications of quantile regression over the past forty years which was further extended by Tyralis \textit{et al.} \cite{Tyralis2024a}.


\subsection{Extreme value theory}
Extreme value theory (EVT) \cite{mcneil1999extreme, pickands1975statistical} forms the basis of extreme value prediction which focuses on modelling the probability distribution of the tail of the data. It provides a set of methods and distribution models for modeling and analysing extreme events. The generalised extreme value distribution (GEV) \cite{bali2003generalized} and the generalised Pareto distribution (GPD) \cite{castillo1997fitting} are two important tools to define and estimate models in EVT. The block maxima (BM) \cite{Ferreira2015on} method and the Peak Over Threshold (POT) \cite{mcneil1999extreme} method are two common model parameter estimation methods for extreme value analysis.

The application of the two main theorems of extreme value theory includes (i) the main limit theorem of EVT (proved by Fréchet \cite{Frechet1927probabilite} in 1927 for the Pareto-type limit distribution and by Fisher and Tippet \cite{Fisher1928limiting} in 1928 for the Weibull and Gumbel limit distributions) leading to the GEV distribution, and (ii) the Gnedenko–Pickands–Balkema–de Haan theorem \cite{Pisarenko2014characterization} leading to the GPD distribution. In the case when the extreme value distribution of a random variable meets certain conditions, exceedance over a high threshold can be approximated by the GPD, while the GEV distribution can approximate the block maxima. The GEV distribution is used in the block BM method for analysing extremes \cite{Ferreira2015on}, while the GPD distribution is used in the POT method for threshold exceedances analysis \cite{mcneil1999extreme}. Comparing the BM method with the POT method \cite{Bücher2021a}, POT is more suitable for quantile estimation as it can better utilise extreme observations with a larger sample size, whilst BM is more suitable for estimating the return level, which refers to the threshold value that is expected to be exceeded once within a particular time period. 

There are various methods for extreme value prediction, including parametric \cite{soukissian2015effect}, non-parametric \cite{schaumburg2012predicting}, and semi-parametric approaches \cite{nascimento2012semiparametric}. Parametric methods assume that the data follows a specified extreme value distribution and makes predictions by estimating parameters. Non-parametric methods use the data directly for prediction and do not rely on a specific distribution structure. Semi-parametric methods combine the advantages of both approaches, utilising data characteristics whilst also considering specific distributions.

\subsection{Multi-step time series prediction}

Single-step prediction refers to a model prediction one step ahead in time, while multi-step time series is a task that aims to predict multiple time steps into the future  \cite{Sandya2013feature}. This becomes increasingly complex with the number of forecast steps, particularly with time series data. The strengths and weaknesses of different neural network architectures vary significantly for time series prediction \cite{chandra2021evaluation}.  Since time series prediction depends on temporal patterns, it's essential to carefully select the optimal neural network architecture and training method. The prediction errors can accumulate over time, especially when dealing with chaotic time series datasets. This implies that in order to produce precise results, the predictive capabilities and hyperparameters of different models need to be considered and customised to the given dataset. Chandra \textit{et al.} \cite{chandra2021evaluation} evaluated a variety of deep learning models, including simple recurrent neural networks (RNN), long short-term memory networks (LSTM), bidirectional LSTM networks (BD-LSTM), encoder-decoder LSTM networks (ED-LSTM), and convolutional neural networks (CNN). These models have then been compared on their performance on univariate time series datasets. They reported that bidirectional LSTM and encoder-decoder LSTM networks performed the best in terms of their prediction accuracy which highlights the advantages of utilising deep learning models in handling multi-step ahead time series prediction. 

Chang \textit{et al.} \cite{chang2012reinforced} used real-time recurrent learning  for training RNNs for  flood forecasting, using an iterative approach for  two-step-ahead forecasts. Additionally, Khedkar \textit{et al.} \cite{khedkar2024evaluation}  incorporated EVT into deep learning  models to address extreme flooding issues across Australia's major catchments. They utilised multivariate and multi-step time series prediction and reported that quantile-LSTM outperformed the baseline deep learning models while providing uncertainty estimates in hydrological forecasting.


In recent years, deep learning models have shown considerable potential in predicting cryptocurrency prices, which is an area characterised by high volatility and unpredictability. Bayesian neural networks (BNNs) have been used to deal with volatility \cite{Lampinen2001bayesian}  and uncertainty quantification in predictions by treating model parameters as probability distributions rather than fixed values \cite{Kononenko1989bayesian}. In cryptocurrency price prediction, BNNs provide a way to quantify the uncertainty of predictions \cite{Jang2017empirical} which is particularly important in financial applications that require risk assessment.  Chandra and He \cite{Chandra2021BNN} employed BNNs to investigate the performance of related multi-step-ahead forecasting models for stock prices, during the COVID-19 pandemic and reported that accurate forecasting was challenging due to the high volatility of the stock market. These areas are precisely where BNNs may perform well in volatile markets by providing more reliable uncertainty estimates.
 
 In addition to the use of Bayesian neural networks, Wang \textit{et al.} \cite{wang2023machine} applied machine learning techniques to forecast cryptocurrency volatility utilising intrinsic features (internal and external determinants). Their findings revealed that LSTM networks significantly outperformed traditional volatility models  such as Generalised Autoregressive Conditional Heteroskedasticity (GARCH).
 Wu \textit{et al.} \cite{Wu2024review} evaluated selected deep learning models, including CNNs, Transformer models, and LSTM variants, using datasets from both before and during the COVID-19 pandemic for cryptocurrency price prediction. They emphasised the importance of evaluating models in different scenarios, and identified the convolutional LSTM with a multivariate approach as the most accurate model.

\section{Methodology}

\subsection{Deep learning models}

 RNNs  are  neural networks designed to handle sequential data \cite{article}  which feature recurrent connections   in the hidden layer to represent temporal data \cite{Elman1990}.   RNNs have been extensively utilised for time series forecasting \cite{chandra2021evaluation}. 

 LSTM network \cite{Hochreiter1998} is a variant of an RNN that addresses the problem of   learning of long-term dependencies by conventional RNNs such as the Elman RNN \cite{Elman1990}.   LSTMs are particularly effective for handling temporal data, since they can retain information over longer periods, outperforming conventional  RNNs. LSTM models enhance traditional RNNs by incorporating memory cells that  feature multiple gates to manage information flow. There are 4 components in a LSTM memory cell (unit): the input gate, the forget gate, the output gate and the cell state.   The interaction of these gates is the crucial part, in updating the cell state  which aids in combating issues related to vanishing and exploding gradients \cite{Hochreiter1998} faced by conventional RNNs \cite{article}.


\begin{figure}[htbp!]
    \centering
    \includegraphics[width=10cm]{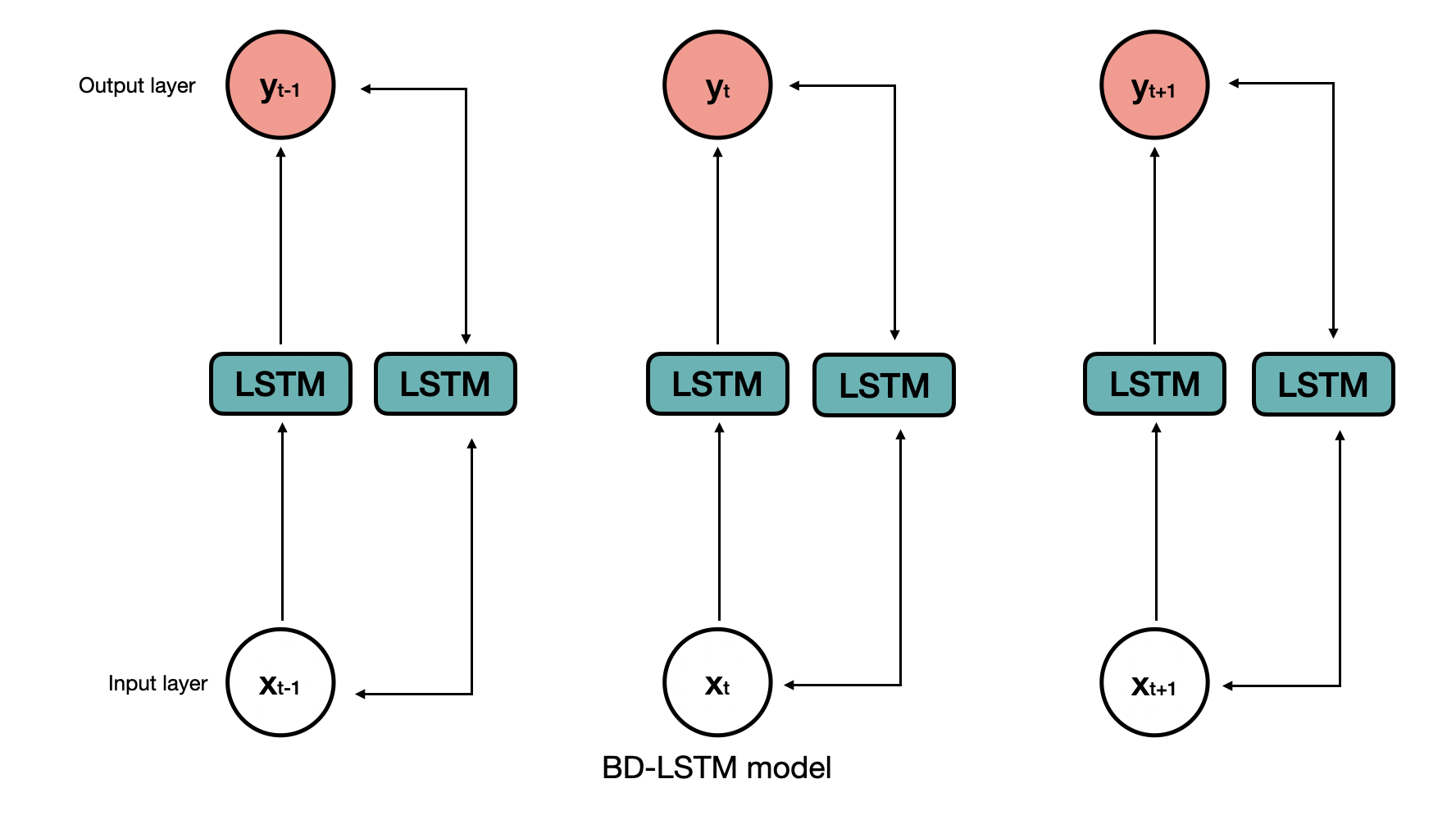}
    \caption{Bidirectional-LSTM network showing the flow of information.}
    \label{fig:BDLSTM}
\end{figure}

The bidirectional long short-term memory (BD-LSTM) is an advanced  LSTM model that handles information in both forward and backward directions through two independent hidden layers, as shown in Figure \ref{fig:BDLSTM}. Unlike canonical LSTM models that process information in a single direction, each input sequence is passed through the RNN twice; once in the forward direction and once in reverse \cite{Graves2005}. This made them prominent for language modelling and natural language processing (NLP) tasks \cite{yu2019review}, and also for multi-step ahead time series forecasting \cite{chandra2021evaluation,Wu2024review}. 
The encoder-decoder long short-term memory (ED-LSTM)  model was designed to handle language modelling tasks \cite{Sutskever2014} which is also effective for time series prediction due to its ability to capture complex temporal patterns and dependencies over long sequences. 
Although the convolutional LSTM (Conv-LSTM) network was initially used for weather forecasting problems \cite{Shi2015}, it is also capable of handling a wide range of time series-related data.  Conv-LSTM can effectively harness both spatial and temporal dependencies in data by combining the strengths of CNNs \cite{alzubaidi2021review} and  LSTM networks. This capability makes Conv-LSTM particularly suited for tasks involving multivariate time series forecasting, such as predicting cryptocurrency prices. Therefore, we used these models for our quantile deep learning framework.

\subsection{Quantile deep learning model}


The key and unique feature of these model implementations is the use of the quantile loss function. For each defined model, there will be a ‘classic’ version with a standard loss function and another version utilising the quantile loss function. This approach allows us to evaluate which set of models performs better, offering more comprehensive predictions that account for the inherent volatility in cryptocurrency markets.

The quantile loss function helps in making predictions that are more tailored to specific sections of the dataset. Instead of predicting only the average outcome, it allows for the prediction of a set of defined quantiles. 
Although the quantile loss function does not predict an exact value, we assume the median values for each time step (prediction horizon)to be the predicted values.

\begin{equation}
    \ell_q (y, \hat{y}) = 
    \begin{cases} 
        q \cdot (y - \hat{y}) & \text{if } y \geq \hat{y} \\
        (q - 1) \cdot (y - \hat{y}) & \text{if } y < \hat{y}
    \end{cases}  
\end{equation}

where, $y$ is the true value, $ \hat{y}$ is the predicted value and $q$ is the specific quantile (e.g. $q=0.95$). We can interpret that if $y \geq \hat{y} $, the actual value is greater than or equal to the predicted value. The loss is given as $q$ times the difference between the true and predicted values. Therefore,  for higher quantiles, the error is higher when the prediction is less than the actual value. In the case that $y < \hat{y}$, the actual value is then less than the predicted value. In this case, the loss is then found to be $(q - 1)$ times the difference between the actual and predicted values. 





\begin{figure}[htbp!]
    \centering
    \subfloat[One-step ahead prediction.]{\includegraphics[width=3in]{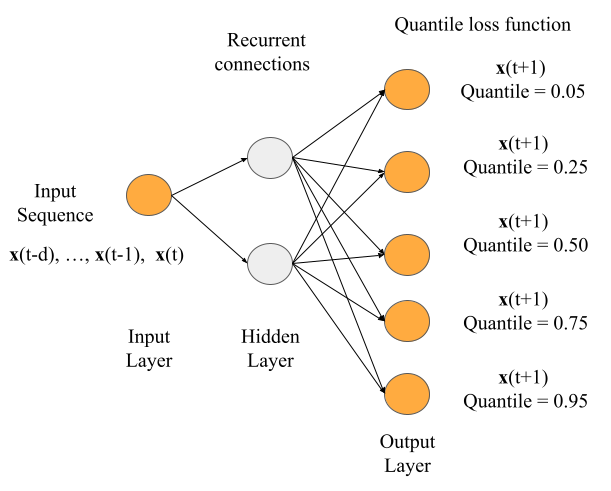}}
 
    \subfloat[Multi-step ahead prediction using grouped quantiles.]{\includegraphics[width=3in]{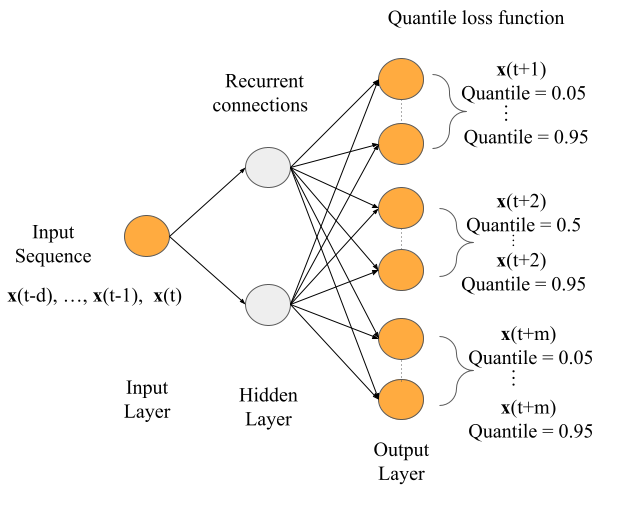}}
    \\
    
    \subfloat[Multi-step ahead prediction using vector-based quantiles for each perdiction horizon.]{\includegraphics[width=3in]{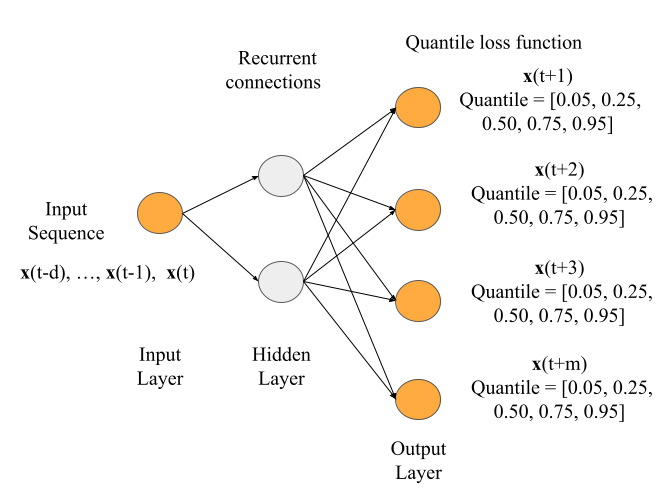}}
    \caption{Quantile recurrent neural network for one-step ahead and multi-step ahead prediction using two strategies, i.) grouped percentiles (Panel b) and ii.) vector-based quantiles (Panel c). Note that the vector-based quantiles have further connections to the hidden neurons, which is not explicitly shown. In the case of the multivariate features, additional neurons in the input layer can be added for each feature. The time-based input and recurrent connections are also not explicitly suing in the recurrent neural network. $x$ represents the time series data index by time $t$ that is windowed by size $d$ for $m$ step-ahead prediction.  }
    \label{fig:quantilenn}
\end{figure}

Applying the quantile loss function to time series data allows for a broader range of predicted values and enables an overview of uncertainties. Instead of predicting a single close price, our implementation will use the quantile loss function that considers a set of quantiles, of a prediction horizon (step). Figure \ref{fig:quantilenn} presents quantile recurrent neural network (RNN) for one-step ahead and multi-step ahead prediction using two strategies, i.) grouped percentiles (Panel b) and ii.) vector-based quantiles (Panel c). $x$ represents the time series data index by time $t$ that is windowed by size $d$ for $m$ step-ahead prediction. Note that the vector-based quantiles have further connections to the hidden neurons, which are not explicitly shown. Furthermore, the time-based input and recurrent connections are also not explicitly shown in the RNN. 
Figure \ref{fig:quantilenn} highlights the interaction between quantile loss function at the output layer of a simple RNN, which is also applicable to other deep learning models (CNN and LSTM models).  We use the quantile loss function instead of the mean squared error loss for the output layer. We are interested to capture the uncertainty in predictions at the 5$^{\text{th}}$, 25$^{\text{th}}$, 50$^{\text{th}}$, 75$^{\text{th}}$ and 95$^{\text{th}}$ percentile, hence use the  quantile values of 0.05, 0.25, 0.5, 0.75 and 0.95.  Note that other quantile values can be defined, as long as it ranges from 0 to 1. After we have defined our quantile values, data is fed into the input layer of the respective neural network model and propagated through the hidden layers, and finally to the output neurons.  This will result in different output and hidden neuron values during back propagations \cite{Hecht-Nielsen1992}. Depending on the quantile value $\tau$, we assign a weight to the quantile loss function and the further $\tau$ deviates from 0.5, the more bias the loss function  producing a lower ($\tau<0.5$) or upper ($\tau>0.5$) value than the median prediction.
 The number of quantile values determines the length of each output neuron. In Figure \ref{fig:quantilenn}, each output neuron features  the predicted values from all the defined quantiles.

In the case of the multivariate features, additional neurons in the input layer can be added for each feature based on Figure \ref{fig:quantilenn}.



\subsection{Framework}


\begin{figure*}[htbp!]
    \centering
    \includegraphics[width=7in]{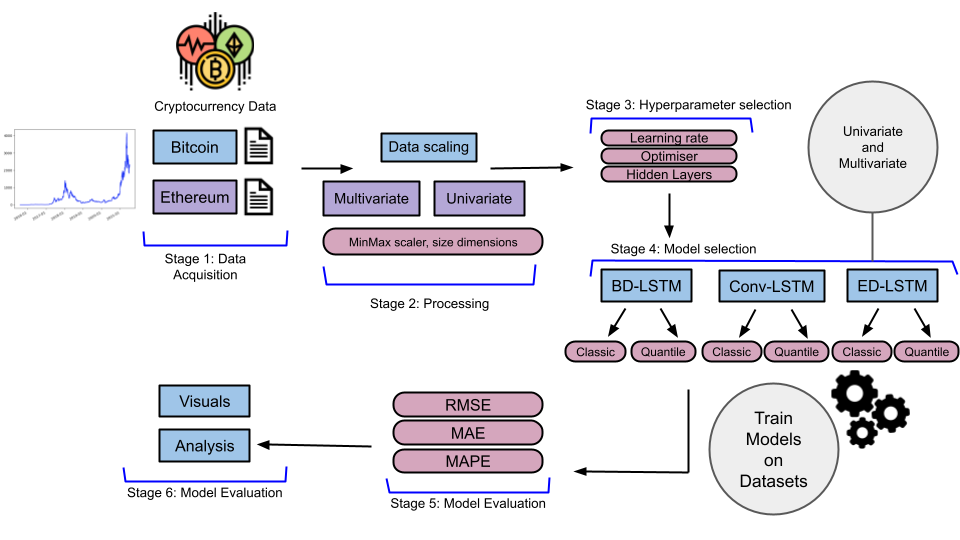}
    \caption{Framework diagram showing the key stages that include data processing model  training and evaluation. We present quartile-based implementation for a  a set of deep learning models including BD-LSTM, Conv-LSTM, and ED-LSTM.}
    \label{fig:framework}
\end{figure*}

The framework presented in Figure \ref{fig:framework} outlines the key components that include data processing and predictions using deep learning models.
In Stage 1, we begin by extracting and processing the selected datasets and applying exploratory data analysis. We need to transform the original time series data into sequences that can be used for prediction.

Stage 2 involves preparing the data for model training.   In the case of deep learning models, we need to process the data depending on their nature, i.e. univariate and multivariate data for associated models as shown in our framework.  This sliding window technique ensures that the model learns from a variety of overlapping sequences, capturing the temporal dependencies in the data. These sequences are then normalised and split into training and testing datasets, as done in previous work in the literature \cite{Wu2024review}.  
 The univariate time series is divided into overlapping windows, each window contains an input sequence vector of a fixed number of consecutive time points (size $d=6$) and an output sequence vector (size $m=5$)  for the future predicted time points.  
 In the case of the multivariate strategy, the model input features include (\textit{high, low, open, close price} and \textit{volume}) to predict the close price for five days (steps). The input features are crucial factors that affect the future close price of the given cryptocurrency, and the previous high and low prices also support estimating the quantiles. In the case of univariate models, we selected close price, as this was determined to be the most important feature in earlier work \cite{Wu2024review}.  
 
In Stage 3, we reviewed the literature to find the most appropriate deep learning models and selected BD-LSTM, Conv-LSTM, and ED-LSTM and defined their hyperparameters  from prior literature \cite{Wu2024review}, to ensure the efficiency of our model. We  define the deep learning model architectures, such as the input size and output size, as shown in Figure \ref{fig:quantilenn}.

In Stage 4, using the three models, we developed a quantile loss function as shown in Figure \ref{fig:quantilenn}.  The most complex part in our framework is setting up and training the multivariate  multi-step ahead quantile-based deep learning models. In both cases, the  multi-step ahead predictions are handled by defining multiple output neurons, with each output neuron representing a distinct step-ahead prediction along with each output neuron presenting a quantile as shown in Figure \ref{fig:quantilenn}-Panel (b) employing the quantile loss function.  We  create both a standard and quantile deep learning model and use the Adam optimiser for training them.


In Stage 5, we   provide analysis of the predictions and review strengths and weaknesses of the respective models and training strategies. We can facilitate a comprehensive comparison of model performance using different metrics including root-mean-square error (RMSE), mean absolute error (MAE), and mean absolute percentage error (MAPE). In our evaluation, we specifically use the RMSE as given below. 
  
\begin{equation}
        \text{RMSE} = \sqrt{\frac{1}{n} \sum_{i=1}^{n} (y_i - \hat{y}_i)^2}  
\end{equation}

where, $n$ is the number of data points (samples), $y$ and $\hat{y}$ are the actual and predicted values, respectively. In the case of multi-step ahead prediction, we take the mean of the RMSE of the $m$ steps. We report RMSE of the different quantiles, where the $\hat{y}$ is of a specific quantile (e.g. quantile value of  0.95).

\subsection{Data}

We demonstrate the effectiveness of quantile deep learning models using Multivariate and Univariate time series datasets such as cryptocurrency using processed data taken from \cite{Wu2024review} (Bitcoin and Ethereum)  and Sunspot, Mackey-Glass and Lorenz time series from Chandra et al. \cite{chandra2021evaluation}.  
\begin{figure*}[h]
    \centering
    \subfloat[Bitcoin]{
        \includegraphics[width=0.50\textwidth]{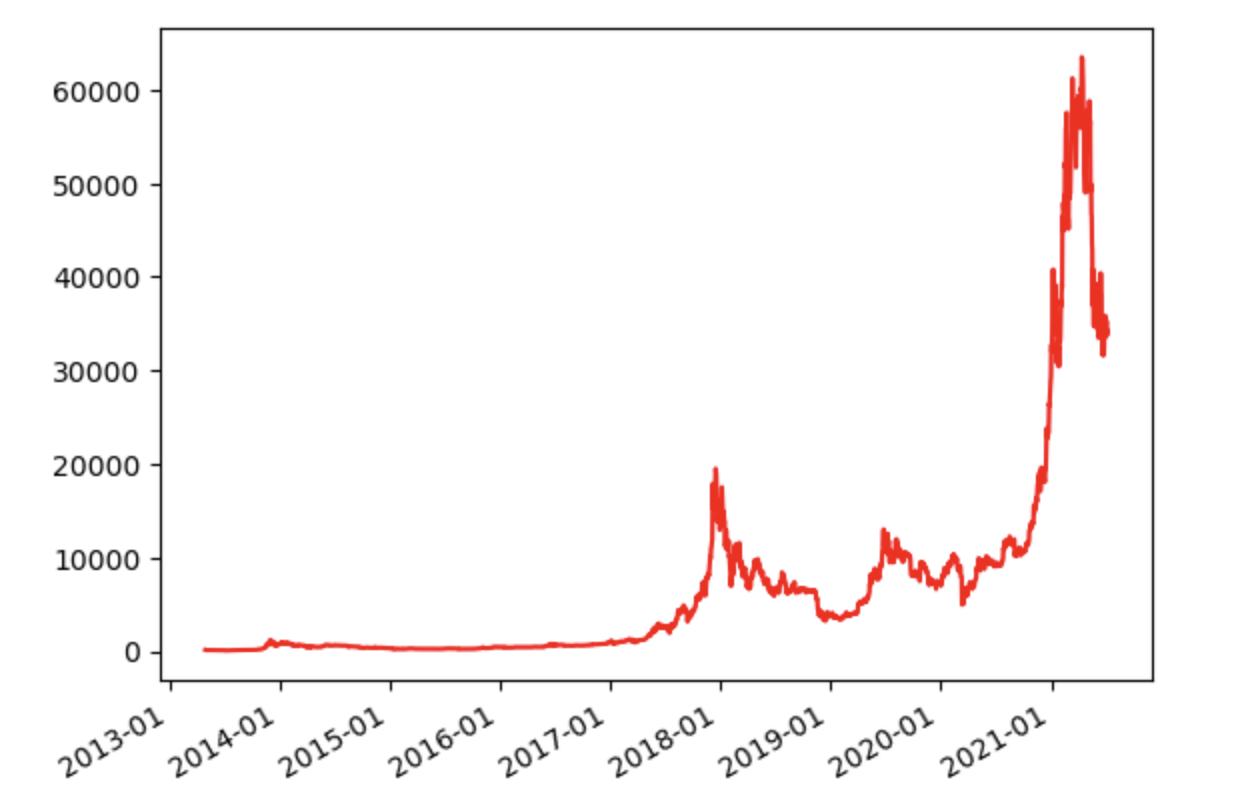}
    }
    \subfloat[Ethereum]{
        \includegraphics[width=0.5\textwidth]{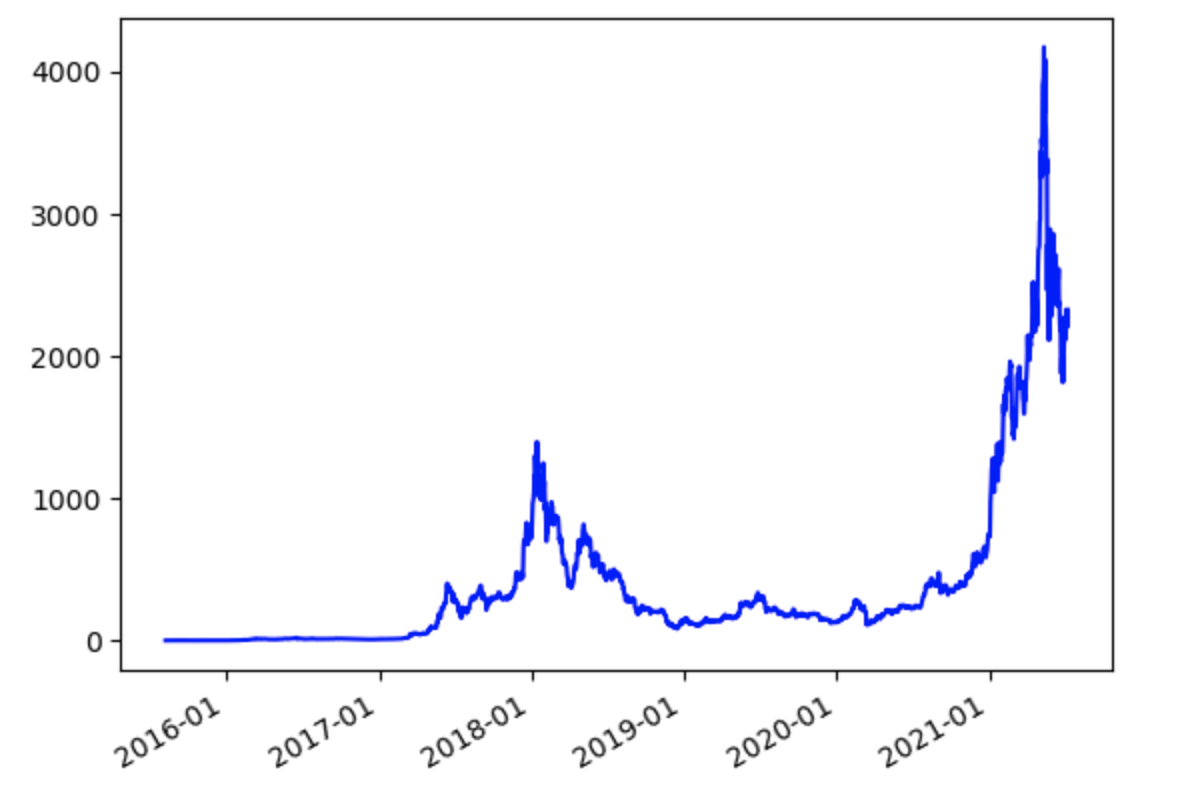}
    } \\
    \caption{Cryptocurrency time series reporting daily close prices}
    \label{U-RMSEstep}
\end{figure*}

\begin{enumerate} 
       \item Bitcoin is a Multivariate dataset that contains daily entries of Bitcoin prices - high, price low, open and close prices - along with trade volume and market capitalisation. There are 2991 daily observations, dating from April 2013 to July 2021. 
        \item Ethereum  is a Multivariate dataset that contains daily entries of Ethereum prices - high, price low, open and close price - along with trade volume and market capitalisation. There are 2160 daily observations dating from August 2015 to July 2021. Both cryptocurrency datasets contains columns such as serial number, name, symbol and date but we will omit them as we won't be needing those in our models.

        \item Sunspot is a univariate dataset that records monthly observations of the sun's surface dating from 1749 to 2021 where the number of sunspots fluctuates and follows an approximate 11-year cycle.  
         
        \item The Mackey-Glass is a univariate dataset that features a continuous chaotic time series, computed with the following delayed differential equation \cite{ReservoirPy2021}.   In this study,  we  use Mackey-Glass \cite{MackeyGlass1977} parameters where $a = 0.2$, $b = 0.1$ and $\tau = 10$. We have generated 3000 time steps and have set a seed for data reproducibility.

     \item     The Lorenz equations \cite{Lorenz1960} three-dimensional chaotic time series composed of ordinary differential equations,  inherently unpredictable over long periods. We have used the default values of the Lorenz system, where  $\rho = 28$, $\sigma = 10$ and $\beta = 2.667$. We generated 10000 time steps and have partitioned the dataset into three univariate time series.

\end{enumerate}

\subsection{Experiment setup}

After developing the initial models, we considered several factors for selecting the appropriate hyperparameters for each model type. 
 Since Bitcoin and Ethereum are highly volatile, training on continuous data would not adequately prepare the model for handling such fluctuations. Therefore, we created the training dataset using a split that was randomly selected, i.e. 80:20 ratio.  The reason for the random train test split is to ensure the models account for data across all time periods. For instance, cryptocurrency data is especially volatile during the COVID-19 pandemic period, which falls only in the test dataset if the train test split wasn't implemented. Our goal  is to ensure that the respective models have the ability to manage volatile data effectively.

We kept the models consistent with previous work (\cite{chandra2021evaluation}) and hence used the hyperparameters presented in Table \ref{tab:hp}.  In the respective deep learning models,  we use adaptive moment estimation (Adam) \cite{kingma2017} optimiser for training with a learning rate of 0.0001.  
In the case of the cryptocurrency datasets (Bitcoin and Ethereum), we use 6 as the input window  size with  5 outputs  (5 prediction horizons) as done by Wu \textit{et al.} \cite{Wu2024review}. In other real-world and simulated time series datasets, the input and output window sizes were adjusted to allow comparison with related work by Chandra \textit{et al.} \cite{chandra2021evaluation}, where the input window is fixed at 5, and the output window at 10 (10 prediction horizons). Furthermore, the following needs to be taken into account along with information in Table \ref{tab:hp}. 
 
\begin{itemize}
 \item  The BD-LSTM model includes both the forward and backward LSTM layer.

 \item  ED-LSTM includes two LSTM networks with a time distributed layer, in the Encoder and Decoder submodels. 

 \item  The Conv-LSTM includes a  1D convolutional layer for the univariate time series and the 2D layer for the multivariate time series.  In the convolutional layer, \textcolor{black}{we use 64 filters with a kernel size of 2.} It also utilises LSTM network and a dense layer.

we use 64 filters with a kernel size of 2

\end{itemize}

\begin{table*}[htbp!]
\centering 
\small

\begin{tabular}{l l l l l l l }
\hline
\hline
 Model & Strategy  &  Input &  Hidden Layers  & \textbf{Output} &   \\
\hline
BD-LSTM & Univariate & [$f=1$, $d=6$]  & [$h_2=50$, $h_1=50$] & 5   \\ 
 & Multivariate&[$f=6$, $d=6$] & [$h_1=50$, $h_2=50$]  & 5    \\ 
ED-LSTM &  Univariate & [$f=1$, $d=6$]  & [$h_1=100$, $h_2=100$]  & 5  \\ 
 &  Multivariate& [$f=6$, $d=6$] & [$h_1=100$, $h_2=100$]  & 5    \\ 
Conv-LSTM & Univariate & [$f=1$, $d=6$]  & [$h_2=20$, $h_1=20$] & 5    \\

 & Multivariate& [$f=6$, $d=6$] & [$h_1=20$, $h_2=20$] & 5     \\
\hline
\hline
\end{tabular}
\caption{Model architecture for univariate/multivariate strategy for the respective deep learning models for the cryptocurrency datasets. We  present number of neurons in input layer (number of features $f$ and window size $d$), hidden layers $h_1,h_2$, and output layer.} Note that number of neurons in input layer and output layer varies in the rest of the datasets. 
\label{tab:hp}
\end{table*}

We report the RMSE mean and 95\% confidence intervals from the test dataset based on 30 independent experimental runs. We note that a lower RMSE indicates better model performance and high uncertainty is indicated by a high confidence interval. In the case of our quantile-based deep learning models, we calculate the average RMSE across the number of time steps at each quantile (0.05, 0.25, 0.5, 0.75, and 0.95), with the mean representing the median value (0.5).

\section{Results}

As outlined earlier, we develop quantile deep learning models for time series prediction including BD-LSTM, ED-LSTM, and Conv-LSTM as the base models and their corresponding quantile versions (e.g. Quantile BD-LSTM).

\subsection{Cryptocurrency datasets}

\begin{figure*}[htbp!]
\subfloat[Univariate strategy]{ 
    \includegraphics[width=0.49\textwidth]{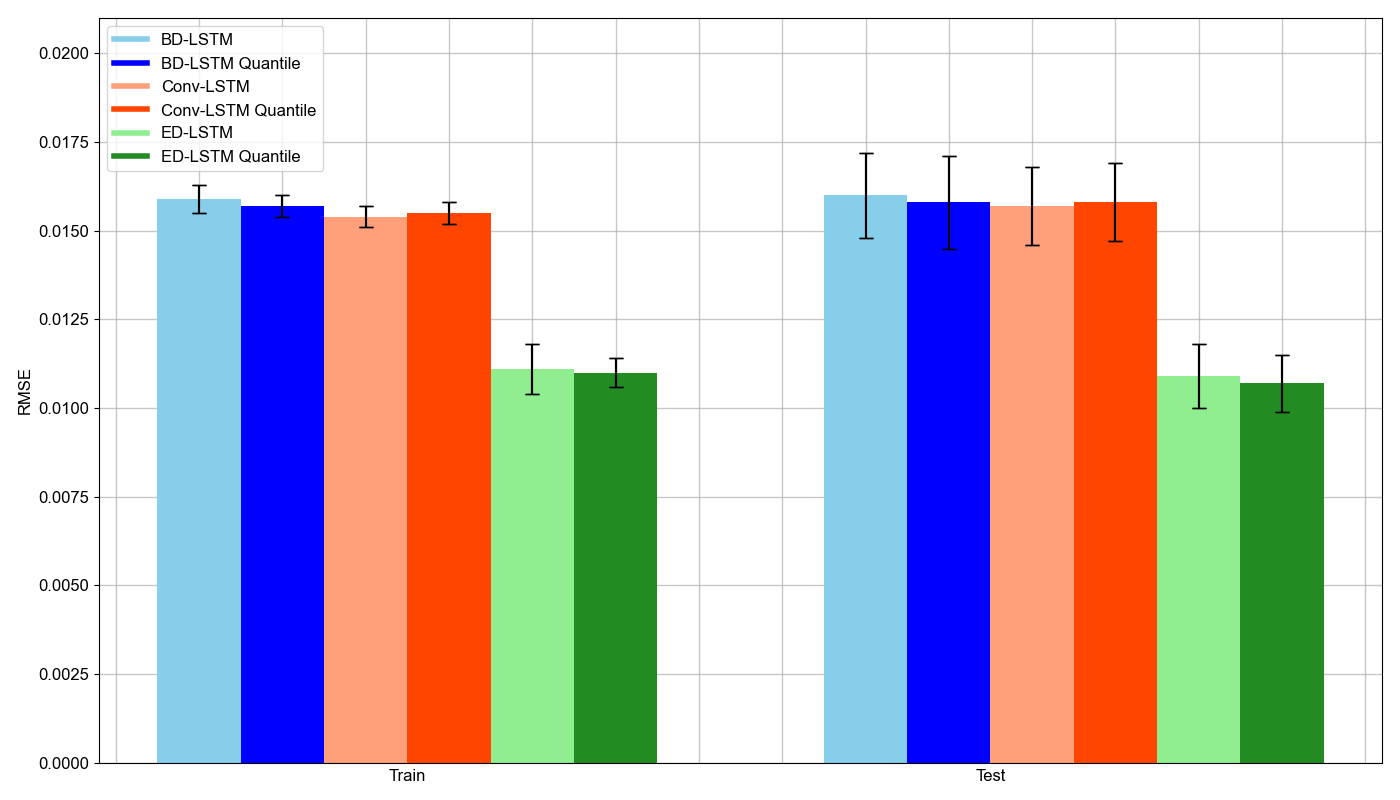}}
\subfloat[Univariate strategy]{ 
    \includegraphics[width=0.49\textwidth]{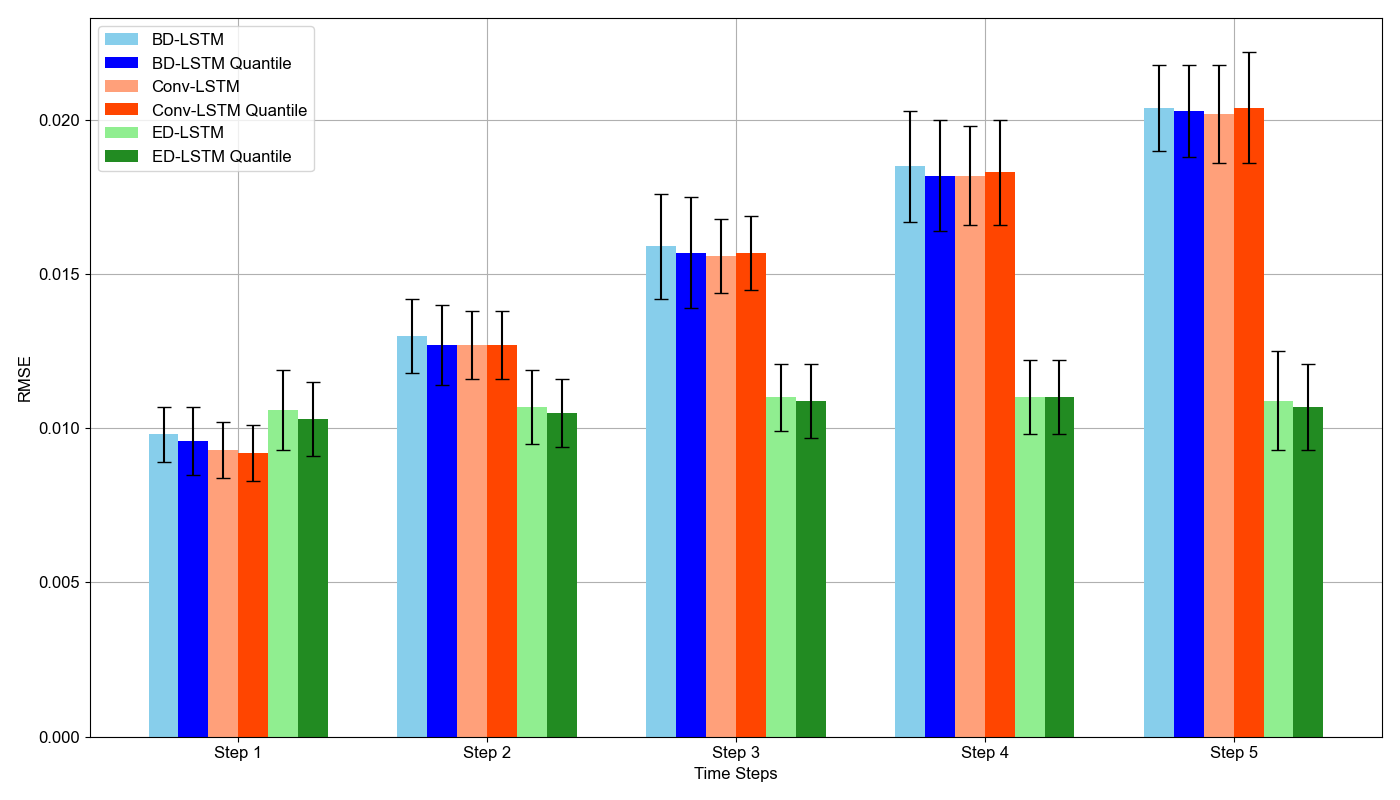}}\\ 
\subfloat[Multivariate strategy]{ 
    \includegraphics[width=0.49\textwidth]{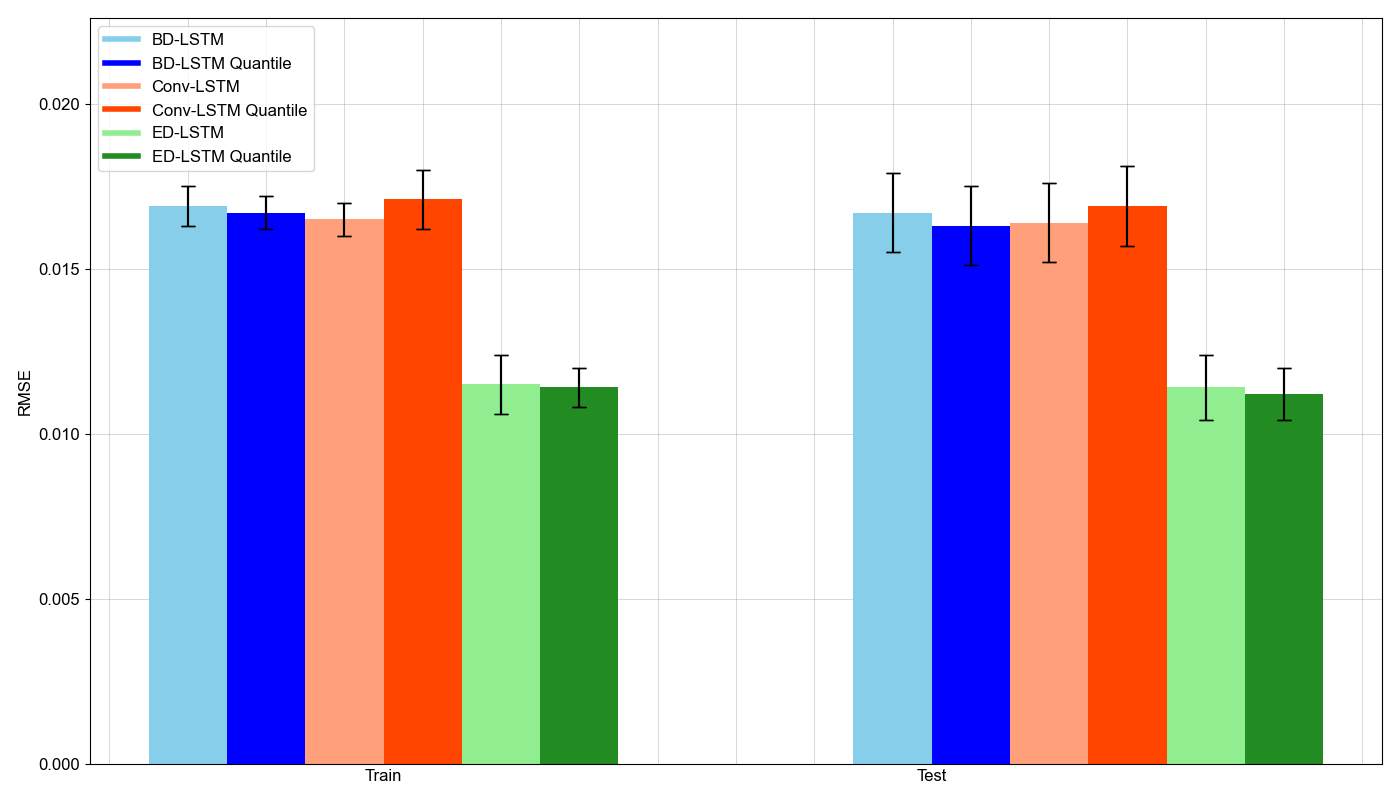}}
\subfloat[Multivariate strategy]{ 
    \includegraphics[width=0.49\textwidth]{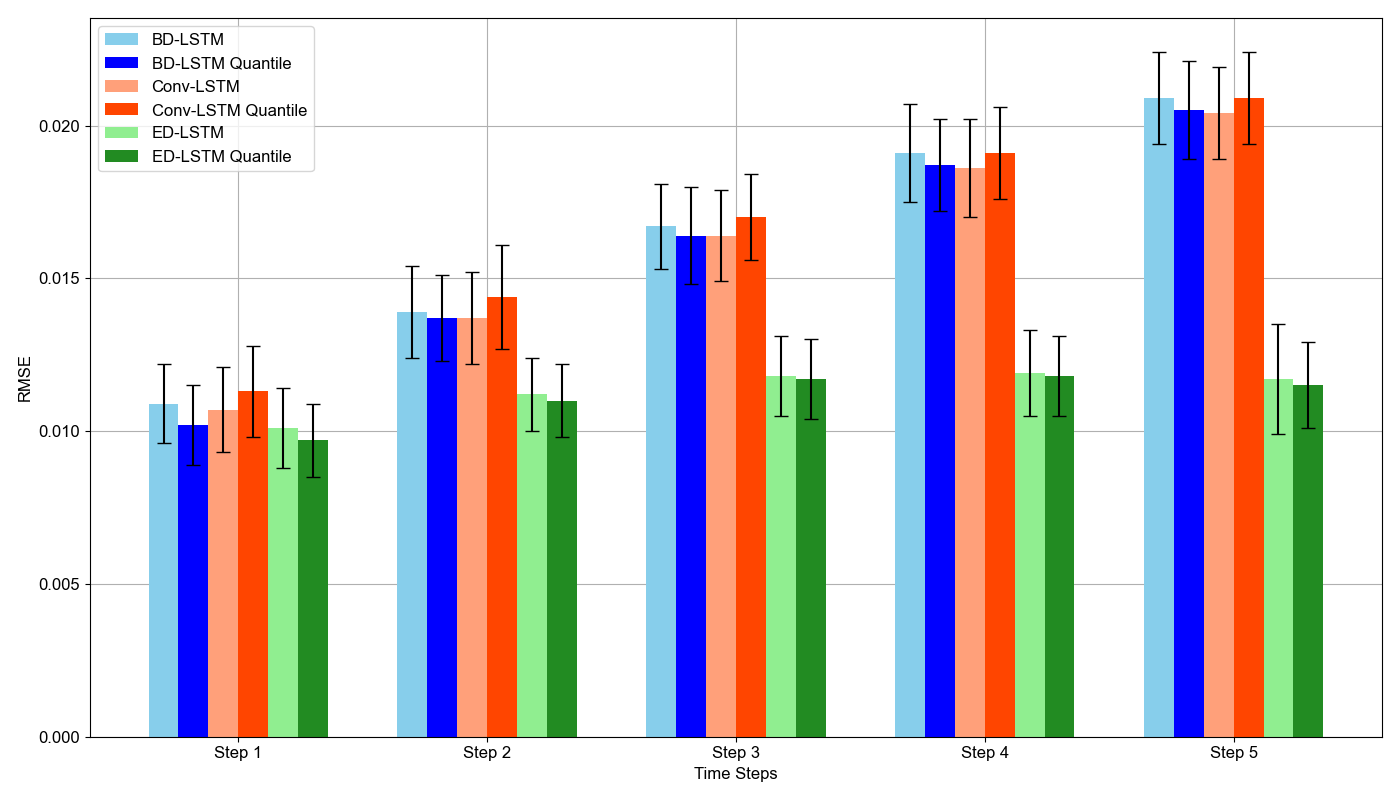}
}
\caption{Bitcoin time series: prediction plots of respective univariate and multivariate strategies (RMSE mean with 95\% confidence interval given as error bar).}
\label{fig:bitcoin}
\end{figure*}
 
\begin{table*}[htbp!]
\centering
\small
\begin{tabular}{c c c c c c c}
    \hline
    \hline
    \multicolumn{7}{c}{\small \textbf{Linear regression}} \\
    \hline
    \textbf{Model} & \textbf{Mean} & \textbf{Step 1} & \textbf{Step 2} & \textbf{Step 3} & \textbf{Step 4} & \textbf{Step 5} \\
    \hline
    Univariate & 0.0062 $\pm$ 0.0058 & 0.0068 $\pm$ 0.0105 & 0.0076 $\pm$ 0.0185 & 0.0076 $\pm$ 0.0149 & 0.0062 $\pm$ 0.0110 & 0.0029 $\pm$ 0.0040 \\
    Quantile Univariate & \textbf{0.0059 $\pm$ 0.0055*} & 0.0067 $\pm$ 0.0103 & \textbf{0.0070 $\pm$ 0.0172} & 0.0075 $\pm$ 0.0152 & \textbf{0.0058 $\pm$ 0.0099} & 0.0027 $\pm$ 0.0044 \\
    Multivariate & 0.0061 $\pm$ 0.0056 & \textbf{0.0066 $\pm$ 0.0101} & 0.0079 $\pm$ 0.0185 & \textbf{0.0069 $\pm$ 0.0126} & 0.0064 $\pm$ 0.0123 & \textbf{0.0027 $\pm$ 0.0042} \\
    Quantile Multivariate & 0.0158 $\pm$ 0.0021* & 0.0092 $\pm$ 0.0011 & 0.0129 $\pm$ 0.0013 & 0.0163 $\pm$ 0.0017 & 0.0193 $\pm$ 0.0017 & 0.0215 $\pm$ 0.0022 \\
    \hline
    \multicolumn{7}{c}{\small \textbf{Univariate deep learning models}} \\
    \hline
    \textbf{Model} & \textbf{Mean} & \textbf{Step 1} & \textbf{Step 2} & \textbf{Step 3} & \textbf{Step 4} & \textbf{Step 5} \\
    \hline
    BD-LSTM & 0.0155 $\pm$ 0.0018 & 0.0098 $\pm$ 0.0009 & 0.0130 $\pm$ 0.0012 & 0.0159 $\pm$ 0.0017 & 0.0185 $\pm$ 0.0018 & 0.0204 $\pm$ 0.0014 \\
    Quantile BD-LSTM & 0.0153 $\pm$ 0.0018* & 0.0096 $\pm$ 0.0011 & 0.0127 $\pm$ 0.0013 & 0.0157 $\pm$ 0.0018 & 0.0182 $\pm$ 0.0018 & 0.0203 $\pm$ 0.0015 \\
    Conv-LSTM & 0.0152 $\pm$ 0.0018 & 0.0093 $\pm$ 0.0009 & 0.0127 $\pm$ 0.0011 & 0.0156 $\pm$ 0.0012 & 0.0182 $\pm$ 0.0016 & 0.0202 $\pm$ 0.0016 \\
    Quantile Conv-LSTM & 0.0153 $\pm$ 0.0019* & \textbf{0.0092 $\pm$ 0.0009} & 0.0127 $\pm$ 0.0011 & 0.0157 $\pm$ 0.0012 & 0.0183 $\pm$ 0.0017 & 0.0204 $\pm$ 0.0018 \\
    ED-LSTM & 0.0108 $\pm$ 0.0006 & 0.0106 $\pm$ 0.0013 & 0.0107 $\pm$ 0.0012 & 0.0110 $\pm$ 0.0011 & \textbf{0.0110 $\pm$ 0.0012} & 0.0109 $\pm$ 0.0016 \\
    Quantile ED-LSTM & \textbf{0.0107 $\pm$ 0.0006*} & 0.0103 $\pm$ 0.0012 & \textbf{0.0105 $\pm$ 0.0011} & \textbf{0.0109 $\pm$ 0.0012} & \textbf{0.0110 $\pm$ 0.0012} & \textbf{0.0107 $\pm$ 0.0014} \\
    \hline
    \multicolumn{7}{c}{\small \textbf{Multivariate deep learning models}} \\
    \hline
    \textbf{Model} & \textbf{Mean} & \textbf{Step 1} & \textbf{Step 2} & \textbf{Step 3} & \textbf{Step 4} & \textbf{Step 5} \\
    \hline
    BD-LSTM & 0.0163 $\pm$ 0.0017 & 0.0109 $\pm$ 0.0013 & 0.0139 $\pm$ 0.0015 & 0.0167 $\pm$ 0.0014 & 0.0191 $\pm$ 0.0016 & 0.0209 $\pm$ 0.0015 \\
    Quantile BD-LSTM & 0.0159 $\pm$ 0.0018* & 0.0102 $\pm$ 0.0013 & 0.0137 $\pm$ 0.0014 & 0.0164 $\pm$ 0.0016 & 0.0187 $\pm$ 0.0015 & 0.0205 $\pm$ 0.0016 \\
    Conv-LSTM & 0.0160 $\pm$ 0.0017 & 0.0107 $\pm$ 0.0014 & 0.0137 $\pm$ 0.0015 & 0.0164 $\pm$ 0.0015 & 0.0186 $\pm$ 0.0016 & 0.0204 $\pm$ 0.0015 \\
    Quantile Conv-LSTM & 0.0165 $\pm$ 0.0017* & 0.0113 $\pm$ 0.0015 & 0.0144 $\pm$ 0.0017 & 0.0170 $\pm$ 0.0014 & 0.0191 $\pm$ 0.0015 & 0.0209 $\pm$ 0.0015 \\
    ED-LSTM & 0.0113 $\pm$ 0.0007 & \textbf{0.0101 $\pm$ 0.0013} & 0.0112 $\pm$ 0.0012 & 0.0118 $\pm$ 0.0013 & 0.0119 $\pm$ 0.0014 & 0.0117 $\pm$ 0.0018 \\
    Quantile ED-LSTM & \textbf{0.0112 $\pm$ 0.0005*} & 0.0110 $\pm$ 0.0009 & \textbf{0.0110 $\pm$ 0.0011} & \textbf{0.0112 $\pm$ 0.0012} & \textbf{0.0114 $\pm$ 0.0012} & \textbf{0.0113 $\pm$ 0.0013} \\
    \hline
    \hline
\end{tabular}
\caption{  Prediction accuracy on the Bitcoin dataset, reporting accuracy  (mean RMSE   and$\pm$ 95\% confidence interval) for 30 independent model training runs.  
* median across quantile (0.5).}
\label{tab:bitsummary}
\end{table*}

\begin{table*}[htbp!]
    \centering
    \small
    \begin{tabular}{c c c c c c c}
    \hline
    \hline
    \multirow{2}{*}{\textbf{Strategy}} & \multirow{2}{*}{\textbf{Model}} & \multicolumn{5}{c}{\textbf{Quantile}} \\
    & & 0.05 & 0.25 & 0.5 & 0.75 & 0.95 \\
    \hline
    &  Quantile BD-LSTM & 0.0339 $\pm$ 0.0032 & 0.0195 $\pm$ 0.0022 & 0.0158 $\pm$ 0.0013 & 0.0201 $\pm$ 0.0021 & 0.0374 $\pm$ 0.0039 \\
    Univariate & Quantile Conv-LSTM & 0.0307 $\pm$ 0.0022 & 0.0189 $\pm$ 0.0014 & 0.0158 $\pm$ 0.0011 & 0.0190 $\pm$ 0.0015 & 0.0296 $\pm$ 0.0022 \\
    & Quantile ED-LSTM & 0.0272 $\pm$ 0.0032 & 0.0153 $\pm$ 0.0019 & 0.0134 $\pm$ 0.0018 & 0.0161 $\pm$ 0.0026 & 0.0258 $\pm$ 0.0037 \\
    \hline
    &  Quantile BD-LSTM & 0.0313 $\pm$ 0.0033 & 0.0198 $\pm$ 0.0025 & 0.0163 $\pm$ 0.0012 & 0.0197 $\pm$ 0.0019 & 0.0314 $\pm$ 0.0030 \\
    Multivariate & Quantile Conv-LSTM & 0.0322 $\pm$ 0.0038 & 0.0203 $\pm$ 0.0029 & 0.0169 $\pm$ 0.0012 & 0.0201 $\pm$ 0.0021 & 0.0318 $\pm$ 0.0033 \\
    & Quantile ED-LSTM & 0.0226 $\pm$ 0.0028 & 0.0133 $\pm$ 0.0019 & 0.0112 $\pm$ 0.0008 & 0.0131 $\pm$ 0.0014 & 0.0206 $\pm$ 0.0023 \\
    \hline
    \hline
    \end{tabular}
    \caption{\small Bitcoin prediction accuracy  (mean RMSE across 5 time steps) at different quantiles.}
    \label{tab:qbitsummary}
\end{table*}


Table \ref{tab:bitsummary} presents the performance (RMSE) of univariate and multivariate linear regression and deep learning models (BD-LSTM, ED-LSTM, Conv-LSTM) for the Bitcoin test dataset. We highlight in bold the best performance for the respective prediction horizons.  
We observe that the quantile linear regression accuracy (RMSE) is similar to linear regression (mean and prediction horizons given by the steps). This implies that quantile regression can effectively handle the volatility of cryptocurrency data while providing predictions of the respective quantiles, which accounts for uncertainty quantification. 
\textcolor{black}{An interesting observation can be seen for multivariate strategy, where there is a higher mean RMSE but a more condensed confidence interval for multivariate quantile linear model.}


Greaves \textit{et al.} \cite{Greaves2015UsingTB} demonstrated that neural networks are superior model classification than linear regression for Bitcoin price prediction; hence, we move on to deep learning models. Earlier, Wu \textit{et al.} \cite{Wu2024review} showed that the  BD-LSTM, ED-LSTM, and  Conv-LSTM networks provided the best accuracy ranks in the univariate and multivariate strategies for a wider range of deep learning models.  

Across both univariate and multivariate strategies in Table \ref{tab:bitsummary}, the ED-LSTM and quantile ED-LSTM models provide the highest prediction accuracy and consistently the ED-LSTM models outperform BD-LSTM and Conv-LSTM. Specifically, the Quantile-ED-LSTM model provides the best accuracy for all prediction horizons, except step one in the univariate and multivariate strategies. Additionally, in Figure \ref{fig:bitcoin} (b), (d), we can observe that ED-LSTM and Quantile-ED--LSTM both provide consistent accuracy as the prediction horizon changes, thus being the most robust and stable model. We can also note that BD-LSTM and Conv-LSTM provide similar performance, but the Quantile-BD-LSTM model consistently provides higher accuracy than its counterpart (BD-LSTM) across both univariate and multivariate strategies. In Table \ref{tab:qbitsummary}, we can see the accuracy (mean RMSE) for each quantile, not only do quantile models often provide similar predictions, but they also provide further information (quantiles) for uncertainty quantification.


\begin{figure*}[htbp!]
\subfloat[Univariate strategy]{
    \includegraphics[width=0.49\textwidth]{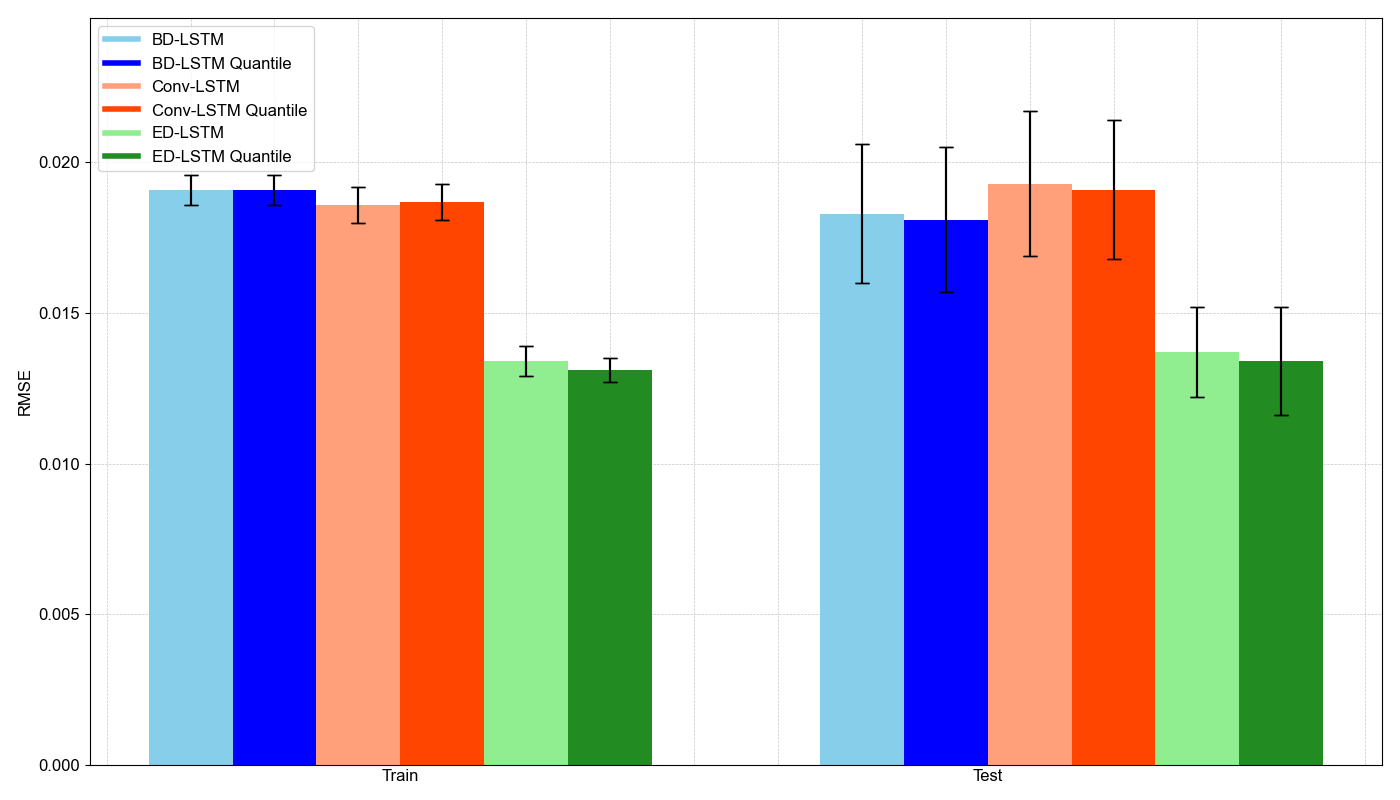}
}
\subfloat[Univariate strategy]{
    \includegraphics[width=0.49\textwidth]{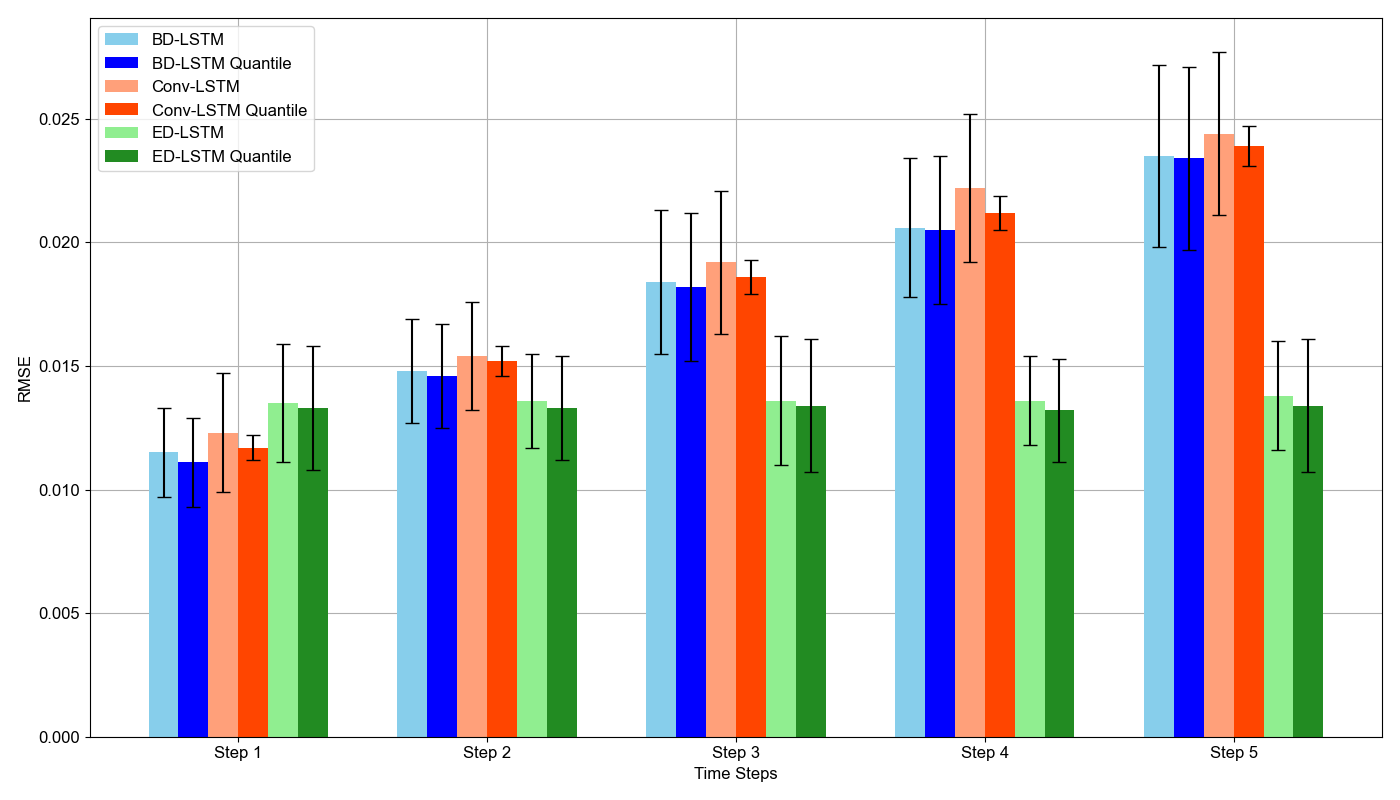}
} \\
\subfloat[Multivariate strategy]{
    \includegraphics[width=0.49\textwidth]{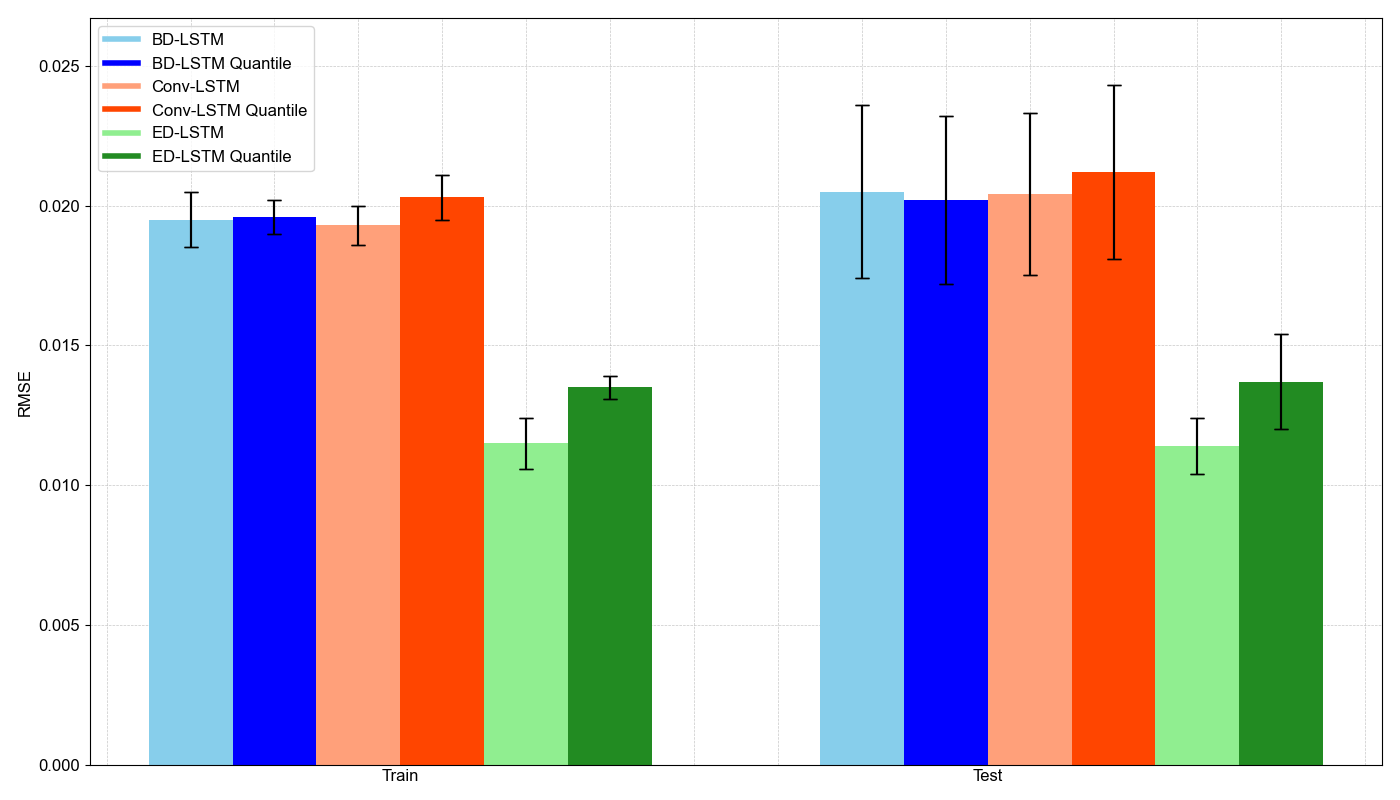}
}
\subfloat[Multivariate strategy]{
    \includegraphics[width=0.49\textwidth]{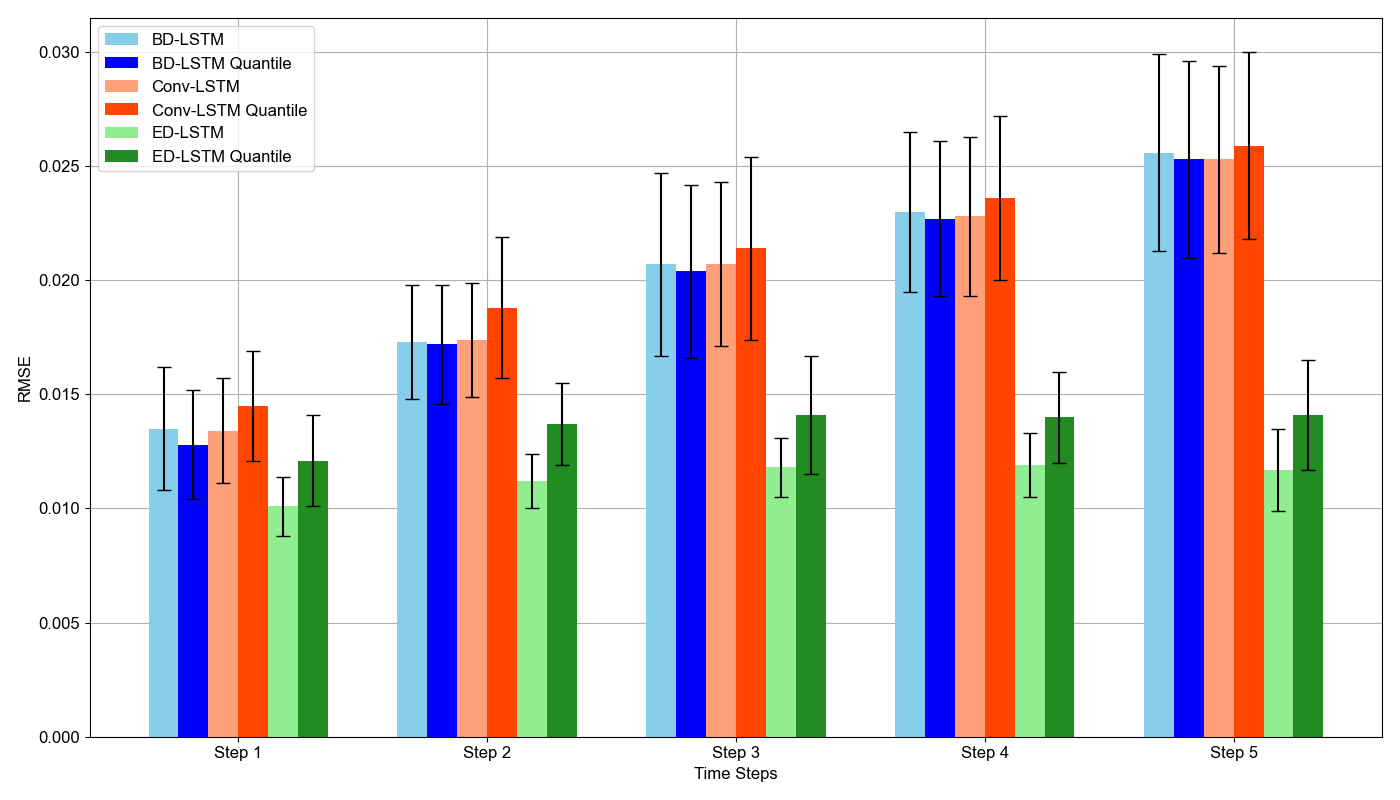}
}
\caption{Ethereum   performance evaluation of respective Univariate and Multivariate deep learning models for 5 prediction horizons (mean RMSE  with 95\% confidence interval as error bar).}
\label{fig:ethereum}
\end{figure*}

In the Ethereum dataset, Table \ref{tab:qethsummary} presents the performance (RMSE) of linear regression and  deep learning models for the test datasets, with the best performance highlighted in bold. We observe that the Univariate quantile models provide the best accuracy (RMSE) and the  Multivariate  quantile models provides the most robustness, as indicated by consistently low confidence intervals. 
Figure \ref{fig:ethpredictions} presents a visualisation of predictions for the respective models and quantiles  for the Ethereum time series, where we observe that the quantiles well capture the actual data points. In Figures \ref{fig:ethereum} (c) and (d), the Multivariate strategy shows that consistently the ED-LSTM models outperform BD-LSTM and Conv-LSTM. In contrast to Bitcoin (Figure \ref{fig:bitcoin}, the classic ED-LSTM model provides the most accurate predictions overall for all time horizons, with the lowest RMSE value. 

\renewcommand{\arraystretch}{1.5}
\begin{table*}[htbp!]
    \centering
    \footnotesize
    \begin{tabular}{c c c c c c c}
        \hline
        \hline
        \multicolumn{7}{c}{\small \textbf{Linear regression}} \\
        \hline
        \textbf{Model} & \textbf{Mean} & \textbf{Step 1} & \textbf{Step 2} & \textbf{Step 3} & \textbf{Step 4} & \textbf{Step 5} \\
        \hline
        Univariate & 0.0079 $\pm$ 0.0060 & 0.0071 $\pm$ 0.0093 & 0.0070 $\pm$ 0.0111 & 0.0092 $\pm$ 0.0126 & 0.0105 $\pm$ 0.0187 & 0.0055 $\pm$ 0.0124 \\
        Quantile Univariate & \textbf{0.0075 $\pm$ 0.0061*} & \textbf{0.0068 $\pm$ 0.0097} & \textbf{0.0067 $\pm$ 0.0111} & \textbf{0.0089 $\pm$ 0.0130} & \textbf{0.0103 $\pm$ 0.0195} & \textbf{0.0050 $\pm$ 0.0126} \\
        Multivariate & 0.0090 $\pm$ 0.0077 & 0.0071 $\pm$ 0.0098 & 0.0090 $\pm$ 0.0149 & 0.0102 $\pm$ 0.0126 & 0.0112 $\pm$ 0.0214 & 0.0074 $\pm$ 0.0228 \\
        Quantile Multivariate & 0.0201 $\pm$ 0.0030* & 0.0115 $\pm$ 0.0025 & 0.0162 $\pm$ 0.0027 & 0.0213 $\pm$ 0.0042 & 0.0242 $\pm$ 0.0039 & 0.0272 $\pm$ 0.0052 \\
        \hline
        \multicolumn{7}{c}{\small \textbf{Univariate deep learning models}} \\
        \hline
        \textbf{Model} & \textbf{Mean} & \textbf{Step 1} & \textbf{Step 2} & \textbf{Step 3} & \textbf{Step 4} & \textbf{Step 5} \\
        \hline
        BD-LSTM & 0.0178 $\pm$ 0.0022 & 0.0115 $\pm$ 0.0018 & 0.0149 $\pm$ 0.0021 & 0.0184 $\pm$ 0.0029 & 0.0206 $\pm$ 0.0028 & 0.0235 $\pm$ 0.0037 \\
        Quantile BD-LSTM & 0.0176 $\pm$ 0.0023* & \textbf{0.0112 $\pm$ 0.0018} & 0.0146 $\pm$ 0.0021 & 0.0182 $\pm$ 0.0030 & 0.0205 $\pm$ 0.0029 & 0.0233 $\pm$ 0.0038 \\
        Conv-LSTM & 0.0187 $\pm$ 0.0023 & 0.0123 $\pm$ 0.0024 & 0.0154 $\pm$ 0.0022 & 0.0193 $\pm$ 0.0029 & 0.0221 $\pm$ 0.0030 & 0.0244 $\pm$ 0.0033 \\
        Quantile Conv-LSTM & 0.0186 $\pm$ 0.0022* & 0.0124 ± 0.0024 & 0.0154 ± 0.0023 & 0.0192 ± 0.0029 & 0.0218 ± 0.0027 & 0.0241 ± 0.0031 \\
        ED-LSTM & 0.0136 $\pm$ 0.0010 & 0.0135 $\pm$ 0.0024 & 0.0136 $\pm$ 0.0019 & 0.0136 $\pm$ 0.0026 & 0.0136 $\pm$ 0.0018 & 0.0138 $\pm$ 0.0022 \\
        Quantile ED-LSTM & \textbf{0.0133 $\pm$ 0.0011*} & 0.0133 $\pm$ 0.0025 & \textbf{0.0133 $\pm$ 0.0021} & \textbf{0.0134 $\pm$ 0.0027} & \textbf{0.0132 $\pm$ 0.0021} & \textbf{0.0134 $\pm$ 0.0027} \\
        \hline
        \multicolumn{7}{c}{\small \textbf{Multivariate deep learning models}} \\
        \hline
        \textbf{Model} & \textbf{Mean} & \textbf{Step 1} & \textbf{Step 2} & \textbf{Step 3} & \textbf{Step 4} & \textbf{Step 5} \\
        \hline
        BD-LSTM & 0.0200 $\pm$ 0.0024 & 0.0134 $\pm$ 0.0024 & 0.0173 $\pm$ 0.0024 & 0.0206 $\pm$ 0.0039 & 0.0229 $\pm$ 0.0034 & 0.0256 $\pm$ 0.0044 \\
        Quantile BD-LSTM & 0.0197 $\pm$ 0.0025* & 0.0128 $\pm$ 0.0025 & 0.0171 $\pm$ 0.0024 & 0.0204 $\pm$ 0.0038 & 0.0228 $\pm$ 0.0035 & 0.0252 $\pm$ 0.0042 \\
        Conv-LSTM & 0.0199 $\pm$ 0.0024 & 0.0134 $\pm$ 0.0023 & 0.0174 $\pm$ 0.0025 & 0.0207 $\pm$ 0.0036 & 0.0228 $\pm$ 0.0035 & 0.0253 $\pm$ 0.0041 \\
        Quantile Conv-LSTM & 0.0208 $\pm$ 0.0024* & 0.0145 $\pm$ 0.0024 & 0.0188 $\pm$ 0.0031 & 0.0214 $\pm$ 0.0040 & 0.0236 $\pm$ 0.0036 & 0.0259 $\pm$ 0.0041 \\
        ED-LSTM & \textbf{0.0113 $\pm$ 0.0007} & \textbf{0.0101 $\pm$ 0.0013} & \textbf{0.0112 $\pm$ 0.0012} & \textbf{0.0118 $\pm$ 0.0013} & \textbf{0.0119 $\pm$ 0.0014} & \textbf{0.0117 $\pm$ 0.0018} \\
        Quantile ED-LSTM & 0.0126 $\pm$ 0.0008* & 0.0121 $\pm$ 0.0020 & 0.0121 $\pm$ 0.0020 & 0.0119 $\pm$ 0.0014 & 0.0132 $\pm$ 0.0016 & 0.0135 $\pm$ 0.0018 \\
        \hline
        \hline
    \end{tabular}
    \caption{Results for the Ethereum dataset reporting accuracy (mean RMSE and $\pm$ 95\% confidence interval) for 30 experimental runs for each model.
  * represents the median across quantiles (0.5). }
  \label{tab:ethsummary}
\end{table*}

Moreover, in Figure \ref{fig:ethereum} (a) and (b), the univariate strategy shows that the ED-LSTM models also consistently outperform BD-LSTM and Conv-LSTM in prediction accuracy with the exception of step one, and Quantile-BD-LSTM provides the best prediction accuracy. Furthermore, the Quantile-ED-LSTM is the most robust univariate model for predicting Bitcoin  as the prediction horizon increases. We note that although BD-LSTM and Conv-LSTM present consistent results,  as the number of prediction days increases, the forecast accuracy gradually decreases. The Quantile-BD-LSTM and Quantile-Conv-LSTM models also consistently provide higher accuracy than their counterparts for the Univariate strategy.  Finally, Conv-LSTM exhibits better performance than the Quantile-Conv-LSTM for multivariate strategies.

\begin{table*}[htbp!]
    \centering
    \small
    \begin{tabular}{c c c c c c c}
    \hline
    \hline
    \multirow{2}{*}{\textbf{Strategy}} & \multirow{2}{*}{\textbf{Model}} & \multicolumn{5}{c}{\textbf{Quantile}} \\
    & & 0.05 & 0.25 & 0.5 & 0.75 & 0.95 \\
    \hline
    & Quantile BD-LSTM & 0.0372 $\pm$ 0.0039 & 0.0219 $\pm$ 0.0025 & 0.0181 $\pm$ 0.0024 & 0.0221 $\pm$ 0.0035 & 0.0390 $\pm$ 0.0045 \\
    Univariate & Quantile Conv-LSTM & 0.0368 $\pm$ 0.0052 & 0.0226 $\pm$ 0.0034 & 0.0191 $\pm$ 0.0023 & 0.0222 $\pm$ 0.0028 & 0.0353 $\pm$ 0.0040 \\
    & Quantile ED-LSTM & 0.0272 $\pm$ 0.0032 & 0.0153 $\pm$ 0.0019 & 0.0134 $\pm$ 0.0018 & 0.0161 $\pm$ 0.0026 & 0.0258 $\pm$ 0.0037 \\
    \hline
    & Quantile BD-LSTM & 0.0385 $\pm$ 0.0042 & 0.0228 $\pm$ 0.0025 & 0.0202 $\pm$ 0.0030 & 0.0249 $\pm$ 0.0053 & 0.0383 $\pm$ 0.0056 \\
    Multivariate & Quantile Conv-LSTM & 0.0392 $\pm$ 0.0041 & 0.0241 $\pm$ 0.0025 & 0.0212 $\pm$ 0.0031 & 0.0267 $\pm$ 0.0057 & 0.0405 $\pm$ 0.0063 \\
    & Quantile ED-LSTM & 0.0265 $\pm$ 0.0031 & 0.0155 $\pm$ 0.0020 & 0.0137 $\pm$ 0.0017 & 0.0168 $\pm$ 0.0029 & 0.0270 $\pm$ 0.0039 \\
    \hline
    \hline
    \end{tabular}
    \caption{Performance evaluation of Multivariate and Univariate strategies for the Ethereum dataset (test dataset mean RMSE across 5-time steps at different quantiles) for 30 independent model training runs.}
    \label{tab:qethsummary}
\end{table*}

  Table \ref{tab:ethsummary} outlines multivariate and univariate strategies for the Ethereum dataset (test dataset mean RMSE across 5 time steps at different quantiles) for 30 independent model training runs. Since the median quantile is our prediction, it has the lowest RMSE compared to any other quantiles. This is logically consistent, as other quantiles cover more extreme prediction values.  
  Additionally, it has a much smaller confidence interval which highlights that quantile models excel in reducing percentage-based errors, making them particularly effective in dealing with price fluctuations and the inherent volatility of cryptocurrency markets. In Figure \ref{fig:ethpredictions}, we present selected predictions for the given quantiles where we observe that in Figure \ref{fig:ethpredictions} (c), the Quantile-Conv-LSTM model performs accurately. Although Conv-LSTM model failed to capture the actual values, its quantile counterpart improved the model's ability to capture the true values.

\subsection{Benchmark datasets}
\begin{figure*}[htbp!]
\centering
\subfloat[Sunspot ]{
    \includegraphics[width=0.49\textwidth]{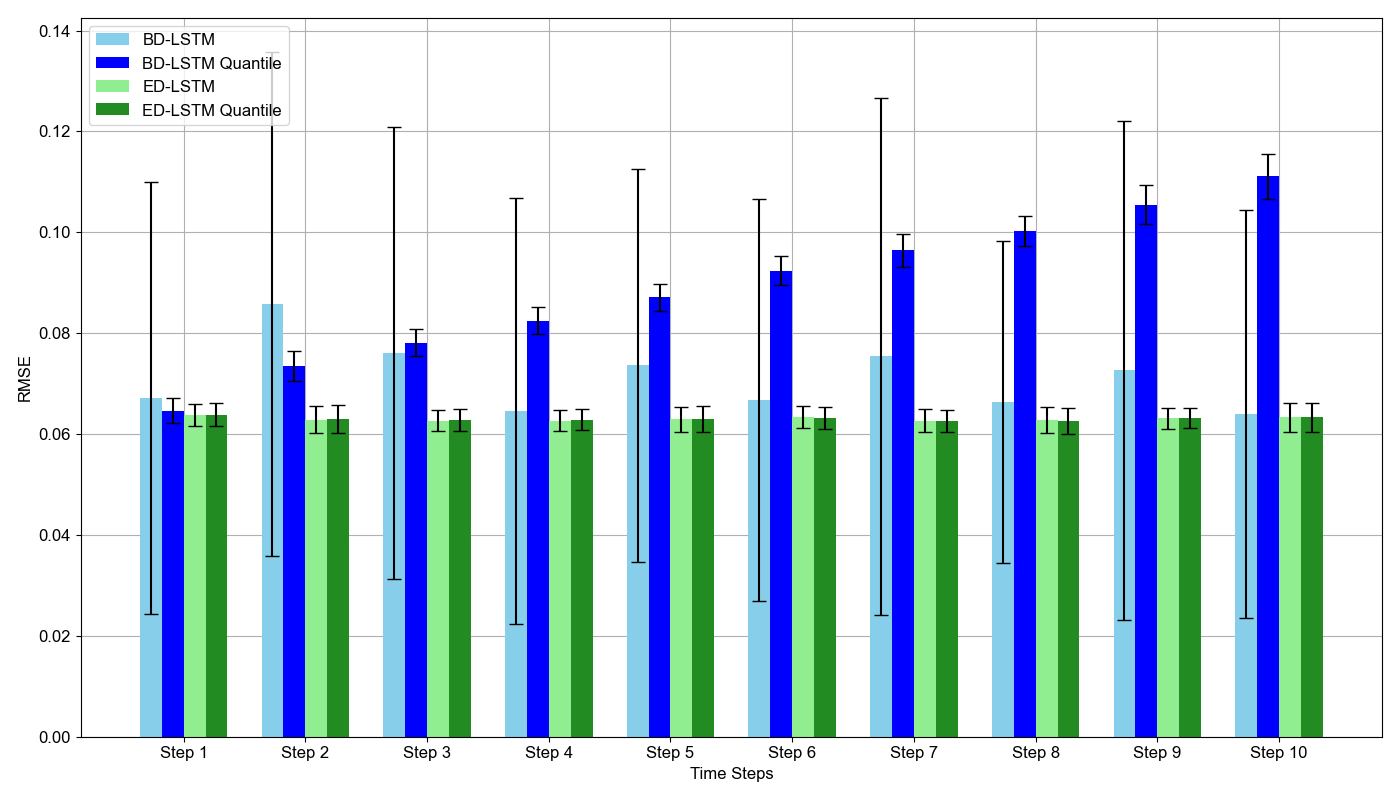}
}
\subfloat[Mackey-Glass 10 step-ahead prediction]{
    \includegraphics[width=0.49\textwidth]{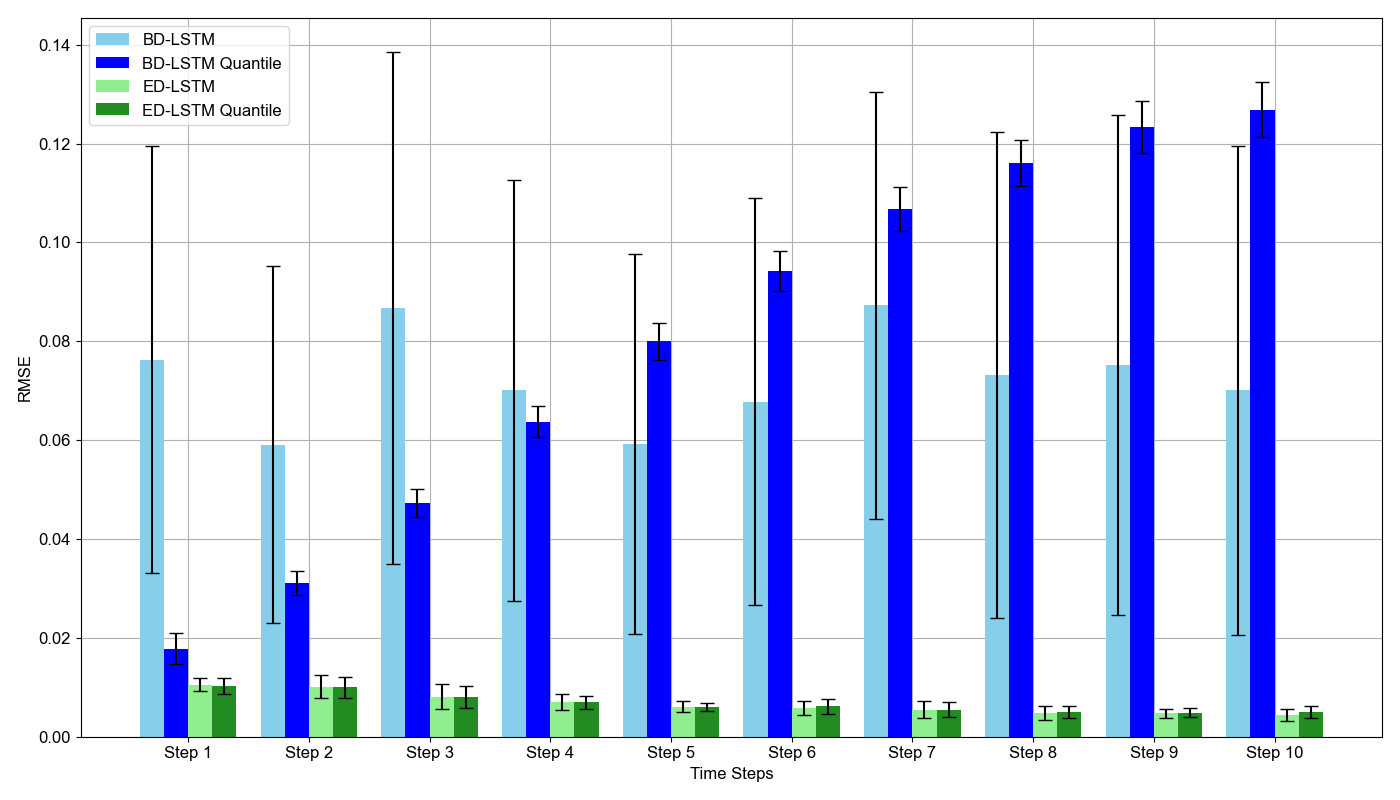}
} \\
\subfloat[Lorenz 10 step-ahead prediction]{
    \includegraphics[width=0.49\textwidth]{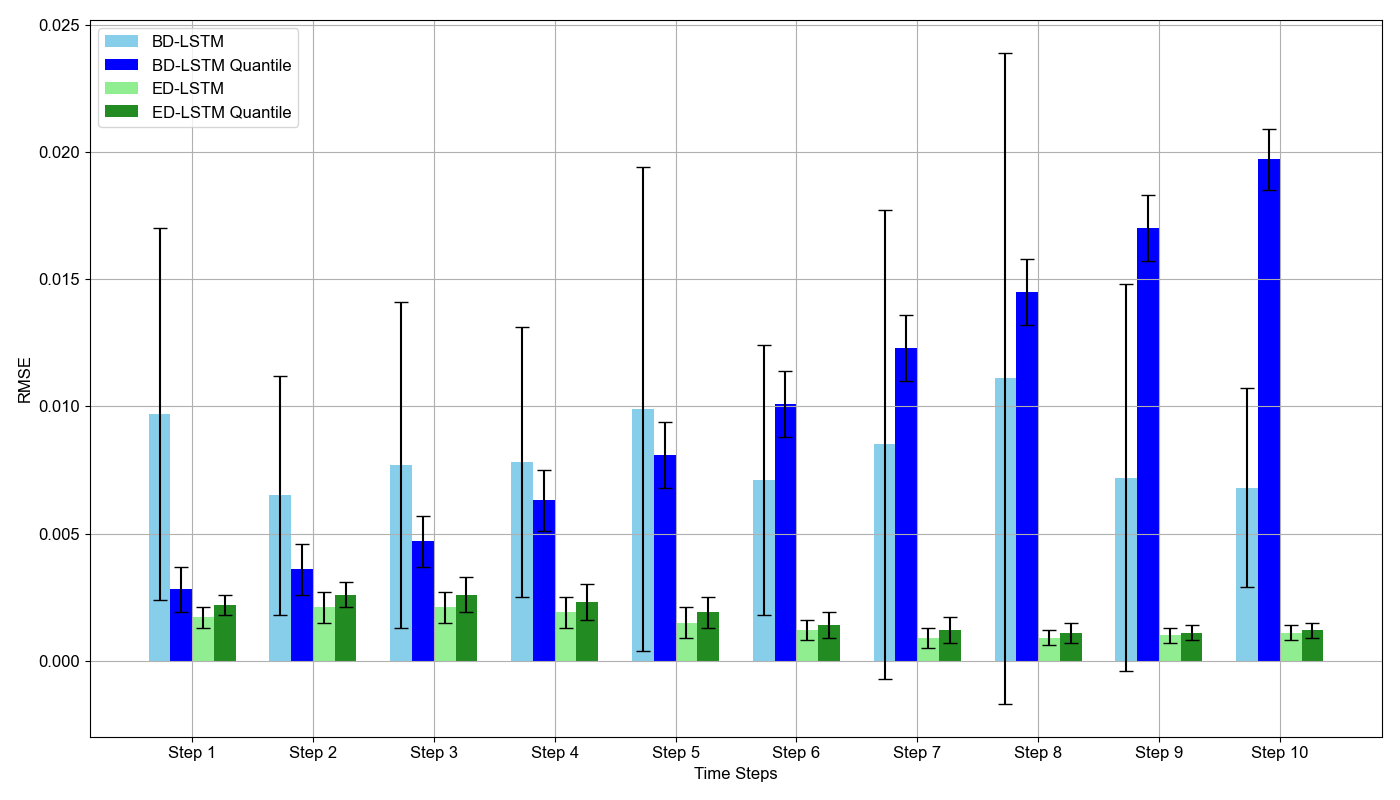}
}
\caption{Performance evaluation of respective Univariate deep learning models, showing 10 step-ahead prediction horizons for benchmark datasets for 30 independent model training runs (mean RMSE with 95\% confidence interval as error bar).}
\label{fig:benchmark10step}
\end{figure*}

Next, we evaluate our framework for the benchmark datasets that include  Sunspots, Mackey-Glass and Lorenz. Chandra \textit{et al.} \cite{chandra2021evaluation} demonstrated that BD-LSTM and ED-LSTM  provided the best accuracy ranks in the evaluation of selected univariate deep learning models. Figure \ref{fig:benchmark10step} presents  10-step prediction horizons of the BD-LSTM, and ED-LSTM models along with their quantile variants. In Figures \ref{fig:benchmark10step} (a)-(c), we observe that the BD-LSTM is the least robust and has the largest error margins (95\% confidence intervals). Additionally, the ED-LSTM and Quantile-ED-LSTM provide a consistent level of accuracy across the prediction horizon, whereas BD-LSTM models are less accurate as time steps increase.
 
Table \ref{tab:benchmark} presents the prediction accuracy for the ED-LSTM and Quantile-ED-LSTM models, both models provided very similar results. In the Sunspot and Lorenz datasets, ED-LSTM models provide the best mean RMSE accuracy and quantile ED-LSTM models exhibit the best performance for the Mackey-Glass time series. We note that there is a distinct difference \textcolor{black}{between ED-LSTM models and the rest of the field, clearly demonstrating that it is the favourable model across various time series datasets.} 
Note that our goal for the Quantile-ED-LSTM is to achieve a similar level of performance to ED-LSTM while providing predictions for the different quantiles as provided. 

\textcolor{black}{We provide prediction visualisations for all datasets in Figure \ref{fig:bitcoipredictions}, \ref{fig:ethpredictions} and \ref{fig:benchmarkpredictions}. All models and their prediction prowess were showcased, often times falling into the quantile range. Note that models tend to perform well for large consistent prices. For low values outputs such as sunspots, mackey-glass and lorenz dataset, we noticed that quantile prediction often deviates from actual values.}

\renewcommand{\arraystretch}{1.5}
\begin{table*}[htbp!]
    \centering
    \footnotesize
    \begin{tabular}{c c c c c c}
    \hline
    \hline
    \multicolumn{6}{c}{\small \textbf{Sunspot}} \\
    \hline
    \textbf{Model} & \textbf{Mean} & \textbf{Step 2} & \textbf{Step 5} & \textbf{Step 8} & \textbf{Step 10} \\
    \hline
    BD-LSTM & 0.0712 ± 0.0139 & 0.0858 ± 0.0499 & 0.0736 ± 0.0389 & 0.0664 ± 0.0319 & 0.0640 ± 0.0405 \\
    Quantile BD-LSTM & 0.0892 ± 0.0045* & 0.0735 ± 0.0029 & 0.0871 ± 0.0027 & 0.1003 ± 0.0030 & 0.1111 ± 0.0045 \\
    ED-LSTM & \textbf{0.0630 ± 0.0008} & \textbf{0.0628 ± 0.0027} & \textbf{0.0629 ± 0.0025} & 0.0627 ± 0.0026 & 0.0633 ± 0.0029 \\
    Quantile ED-LSTM & 0.0630 ± 0.0024* & 0.0629 ± 0.0028 & 0.0630 ± 0.0026 & \textbf{0.0626 ± 0.0026} & \textbf{0.0633 ± 0.0029} \\
    \hline
    \multicolumn{6}{c}{\small \textbf{Mackey-Glass}} \\
    \hline
    \textbf{Model} & \textbf{Mean} & \textbf{Step 2} & \textbf{Step 5} & \textbf{Step 8} & \textbf{Step 10} \\
    \hline
    BD-LSTM & 0.0725 ± 0.0145 & 0.0591 ± 0.0362 & 0.0592 ± 0.0385 & 0.0732 ± 0.0492 & 0.0701 ± 0.0495 \\
    Quantile BD-LSTM & 0.0807 ± 0.0119* & 0.0311 ± 0.0024 & 0.0800 ± 0.0037 & 0.1161 ± 0.0046 & 0.1269 ± 0.0055 \\
    ED-LSTM & 0.0071 ± 0.0011 & 0.0101 ± 0.0023 & 0.0060 ± 0.0011 & \textbf{0.0048 ± 0.0014} & \textbf{0.0049 ± 0.0012} \\
    Quantile ED-LSTM & \textbf{0.0071 ± 0.0010*} & \textbf{0.0100 ± 0.0021} & \textbf{0.0059 ± 0.0008} & 0.0049 ± 0.0012 & 0.0050 ± 0.0012 \\
    \hline
    \multicolumn{6}{c}{\small \textbf{Lorenz}} \\
    \hline
    \textbf{Model} & \textbf{Mean} & \textbf{Step 2} & \textbf{Step 5} & \textbf{Step 8} & \textbf{Step 10} \\
    \hline
    BD-LSTM & 0.0097 ± 0.0017 & 0.0072 ± 0.0053 & 0.0111 ± 0.0097 & 0.0097 ± 0.0086 & 0.0076 ± 0.0055 \\
    Quantile BD-LSTM & 0.0140 ± 0.0013* & 0.0044 ± 0.009 & 0.0104 ± 0.0013 & 0.0177 ± 0.0018 & 0.0239 ± 0.0022 \\
    ED-LSTM & \textbf{0.0015 ± 0.0004} & \textbf{0.0022 ± 0.0007} & \textbf{0.0014 ± 0.0006} & \textbf{0.0009 ± 0.0004} & \textbf{0.0010 ± 0.0004} \\
    Quantile ED-LSTM & 0.0021 ± 0.0004* & 0.0028 ± 0.0005 & 0.0020 ± 0.0007 & 0.0013 ± 0.0004 & 0.0013 ± 0.0004 \\
    \hline
    \hline
    \end{tabular}
    \caption{Benchmark  time series datasets reporting model test accuracy (mean RMSE and $\pm$ 95\% confidence interval for 30 experimental runs) for univariate deep learning models.  
     * represents the median prediction  quantile (0.5).}
     \label{tab:benchmark}
\end{table*}

\section{Discussion}
 
\renewcommand{\arraystretch}{1.2}
\begin{table*}[htbp]
    \centering
    \footnotesize
    \begin{tabular}{m{1.7cm}<{\centering} m{1.3cm}<{\centering} m{1.4cm}<{\centering} m{1.4cm}<{\centering} m{1.6cm}<{\centering} m{1.6cm}<{\centering} m{1.4cm}<{\centering} m{1.4cm}<{\centering}}
    \hline
    \hline
    \textbf{Data} & \textbf{Strategy} & \textbf{BD-LSTM} & \textbf{Quantile BD-LSTM} & \textbf{Conv-LSTM} & \textbf{Quantile Conv-LSTM} & \textbf{ED-LSTM} & \textbf{Quantile ED-LSTM} \\
    \hline
    Bitcoin & \multirow{2}{*}{Univariate} & 6 & 5 & 3 & 4 & 1 & 2  \\
    Ethereum & & 4 & 3 & 6 & 5 & 2 & 1 \\
    \hline
    \multicolumn{2}{c}{Mean Rank} & 5 & 4 & 4.5 & 4.5 & \textbf{1.5} & \textbf{1.5} \\
    \hline
    Bitcoin & \multirow{2}{*}{Multivariate} & 5 & 3 & 4 & 6 & 2 & 1 \\
    Ethereum & & 5 & 3 & 4 & 6 & 2 & 1 \\
    \hline
    \multicolumn{2}{c}{Mean Rank} & 5 & 3 & 4 & 6 & 2 & \textbf{1} \\
    \hline
    \hline
    Sunspot & \multirow{3}{*}{Univariate} & 3 & 4 & - & - & 1 & 2 \\
    Mackey-Glass & & 3 & 4 & - & - & 2 & 1 \\
    Lorenz & & 3 & 4 & - & - & 1 & 2 \\
    \hline
    \multicolumn{2}{c}{Mean Rank} & 3 & 4 & - & - & \textbf{1.33} & 1.67 \\
    \hline
    \hline
    \end{tabular}
    \caption{Performance (rank) of different models for respective time-series problems. Note lower rank denotes better performance.}
    \label{tab:summary}
\end{table*}

This study explored quantile deep learning models for multivariate and multi-step ahead time series prediction with univariate and multivariate models for selected cryptocurrency and time series prediction datasets. Table \ref{tab:summary} presents a summary of the results,   highlighting that both the conventional ED-LSTM and Quantile-ED-LSTM models consistently deliver the strongest predictive performance across all datasets. We can also observe that for the cryptocurrency datasets, neither the BD-LSTM nor the Conv-LSTM  (including their quantile variants) show a clear performance hierarchy. This aligns with the RMSE results outlined in Tables \ref{tab:bitsummary} and \ref{tab:ethsummary}, where all four models exhibit competitive accuracy (close performance). Additionally, the primary goal of this study is to enhance the representation of uncertainty with the quantile loss function, rather than to improve the forecast accuracy of the conventional models. Therefore, we expected a similar performance of quantile models relative to conventional deep learning models,  and has been demonstrated across all datasets in Figures \ref{fig:bitcoin}, \ref{fig:ethereum} and \ref{fig:benchmark10step}.

In both univariate and multivariate cases, the quantile models performed similar to the conventional deep learning models. In the case of Ethereum, \textcolor{black}{the performance for the quantile models are slightly poorer than conventional models (Table \ref{tab:ethsummary}). However, in the Bitcoin dataset (Table \ref{tab:bitsummary}),} the predictions remained similar to the univariate datasets, \textcolor{black}{demonstrating that deep learning models have different predictive abilities depending on the dataset.} We can gather that the model with the most consistent and accurate predictions is the Quantile-ED-LSTM model.
    Furthermore, we can review the results of the other datasets. The best models in the cryptocurrency datasets do not automatically imply they are the best across all other datasets. \textcolor{black}{We ran similar experiments on three other volatile datasets (i.e. sunspots, mackey-glass and lorenz). The BD-LSTM standard model proved to be particularly unreliable, with very large confidence intervals and RMSE values (Figure \ref{fig:benchmark10step}). The BD-LSTM quantile model grew in RMSE value across the time steps but has a small confidence interval, demonstrating the model's robustness. Consistent with the cryptocurrency results, the ED-LSTM models outperformed the BD-LSTM models as seen in Table \ref{tab:benchmark}, where the ED-LSTM models have a higher mean rank than the BD-LSTM models across all datasets. }

We next review the multivariate results for the cryptocurrency datasets \textcolor{black}{(Tables \ref{tab:bitsummary} and \ref{tab:ethsummary}) where the distinguishing features between each model and their quantile counterparts have more clarity and definition. The multivariate results remained consistent with the univariate parts, where the ED-LSTM outperforms both the Conv-LSTM and BD-LSTM models, we can see the contrast between the three models in Figures \ref{fig:bitcoin} and \ref{fig:ethereum}.  }Since ED-LSTM models process the entirety of the given historical data through the encoder and then make predictions through the decoder, they are able to take into account the entire history of the dataset and are hence able to make more valid predictions.  BD-LSTM model also performs poorly in comparison to the ED-LSTM models as seen in Table 7, as they are not as capable in handling long term dependencies in the data, which is a key feature in volatile datasets.



Our findings have better results than those reported in the related study \cite{Wu2024review}, as their test mean for the ED-LSTM multivariate Bitcoin data \textcolor{black}{was 0.0373 compared to our quantile ED-LSTM result of 0.0112 as seen in Table \ref{tab:bitsummary}. However, do note that the train test split ratio are different across the two papers where we used 80:20 in comparison to 70:30 by Wu et al \cite{Wu2024review}. The lower mean RMSE from our ED-LSTM quantile regression analysis particularly emphasises that the median (0.5 quantile) }results yield superior predictive accuracy compared to traditional methods. This enhancement in performance underscores the robustness of quantile regression for time series forecasting. Furthermore, quantile regression not only improves prediction accuracy but also offers a probabilistic interpretation by providing a spectrum of potential outcomes, thereby enriching the decision-making process with a more comprehensive risk assessment. 
In the literature, conventional deep learning models have been evaluated for multi-step ahead time series prediction on Sunspots, Mackey-Glass and Lorenz system \cite{chandra2021evaluation}; where, BD-LSTM had the best accuracy for univariate time series data. Furthermore, Chandra et. al \cite{chandra2021evaluation}; reported a common trend where the predictive accuracy decreases across higher steps-ahead prediction. Note that in their results,  in the case of Mackey-Glass, ED-LSTM was the best-performing model and for  the Lorenz system, both ED-LSTM and LSTM outperformed BD-LSTM.  This is in line with the observations in our study, where ED-LSTM is the best performing model. 

We recall that the process of extreme value forecasting (EVF) \cite{Chou2005, Zhao2014} does not directly calculate extreme values \cite{Dixon1950}, and instead calculates uncertainty bounds \cite{Merz2009}, which in turn, implicitly addresses extreme values. Our quantile deep learning models for multi-step ahead prediction serve as an example of EVF in action, as the quantile nature of the model indicates uncertainty. 

In terms of the limitations, there is   room for improving the hyperparameter tuning \cite{Yang2020hyperparameter}, including adjusting the model topology given by the number of hidden layers, and neurons. \textcolor{black}{We had no indication on how accurate our quantile predictions are. In our paper, we could only mention that the quantile predictions have a higher RMSE value than its median, suggesting them covering the upper and lower range of the prediction. However, there is no set values for us to compare our quantile prediction. }Furthermore, to enhance model accuracy, regularisation techniques such as Lasso \cite{Muthukrishnan2016LASSO} and ridge regression \cite{vanWieringen2015} can be incorporated into our framework. Dropout-based regularisation \cite{Wager2013} has been prominent in deep learning, and this can also be incorporated in the respective model architectures.  Extensive evaluation using different  training-test split ratios, cross-validation, and different input dimension  window can be considered. Finally, model outputs and feature variable assumptions need to be clearly defined and constrained to ensure quantile predictions adequately cover the expected range. \textcolor{black}{Many variables such as price and volume cannot be negative. The output values of other datasets such as sunspots and mackey-glass also has to be positive. Therefore, we should have specified our model predictions to always be greater than 0.}


There are several strategies that can be taken to enhance our quantile deep learning framework further. 
The quantile deep learning model  for predicting cryptocurrency can be further enhanced using multimodal \cite{AlTameemi2023} framework that considers text data such as data from news media and twitter about  cryptocurrency markets. Sentiment analysis using natural language processing and large language models \cite{Zhang2023} can be useful in providing further information for the deep learning models. Sentiment analysis in combination with a quantile deep learning can be very useful in improving future predictions that features the quantiles for robust uncertainty quantification.


There are other approaches to uncertainty quantification, such as Bayesian inference \cite{Neal1996}, where posterior distribution is obtained by prior distribution and likelihood function \cite{Yoon1993}. In particular, uncertainty bound can be calculated by sampling from the posterior distribution  using Markov Chain Monte Carlo (MCMC) \cite{MacKay1992}. There have been efforts in developing Bayesian neural networks and Bayesian deep learning models with MCMC \cite{Zhang2019} and variational-Bayes sampling strategies \cite{Attias2013}. In the field of finance, there exists literature which performs multi-step ahead price forecasting using the Bayesian approach \cite{Chandra2021BNN}. Although our study provides a frequentest approach \cite{Chaput2011} to uncertainty quantification using quantile regression in deep learning, our framework can be extended using Bayesian deep learning. In doing so, we would be sampling from the posterior distribution (model weights and biases) using MCMC or variational Bayes and computing additional uncertainties projected. Although this would be an overkill in conventional problems, it would be useful in problems where risk analysis is vital, such as medical diagnosis. Furthermore, the quantile deep learning model can be utilised for time series data imputation tasks \cite{AfrifaYamoah2020}, where uncertainties obtained from the different quantiles can be useful in producing different versions of impute datasets.

\section{Conclusion}

In this study, we investigated the combination of quantile regression in selected deep learning models for multi-step ahead time series prediction. 
Our results demonstrated that the  combining quantile regression with deep learning models has been very effective, even in volatile environments such as the cryptocurrency markets. In the case of the cryptocurrency datasets, the quantile ED-LSTM model has consistently outperformed traditional methods, highlighting their ability to effectively handle forecast uncertainty and volatility. Although our current models have shown strong predictive capabilities, it is still possible to improve them further through further hyperparameter tuning and incorporating novel architectural and training strategies. 

The overall objective was to not really outperform the existing versions of the models, but rather to see if we can provide more clarity and information with the quantile loss function.  Not only do our quantile models provide exceptional accuracy, but also demonstrate remarkable stability and robustness across various prediction horizons. However, even though our results do show that combining the quantile loss function with deep learning does occasionally provide slightly more accurate predictions. The quantile models offer a more reliable prediction by embracing the inherent uncertainties, providing more information as well as being less sensitive to outliers. This allows us to be more certain within our predictions, making them invaluable tools for navigating datasets that are unpredictable and volatile in  nature.

Our study shows that quantile deep learning models not only improve forecast accuracy but also provide an indication of uncertainty, which is useful in risk assessment and decision-making processes. This process is critical in dealing with high volatility and extreme data in risk-sensitive environments and has potential for modelling climate extreme events. Future research can explore model optimisation and applications to other  forecasting problems and regression tasks.

\section*{Data and Code Availability}

We provide  Python code and the data used in our models using the GitHub repository \footnote{\url{https://github.com/sydney-machine-learning/quantiledeeplearning}}.

\section*{Acknowledgements}

We thank Arpit Kapoor from UNSW Sydney for earlier discussions.

\bibliographystyle{elsarticle-num} 
\bibliography{cas-refs}

\appendix
 \section{Prediction quantiles}

\begin{figure*}[htbp!]
\subfloat[Univariate  Quantile BD-LSTM ]{
    \includegraphics[width=0.49\textwidth]{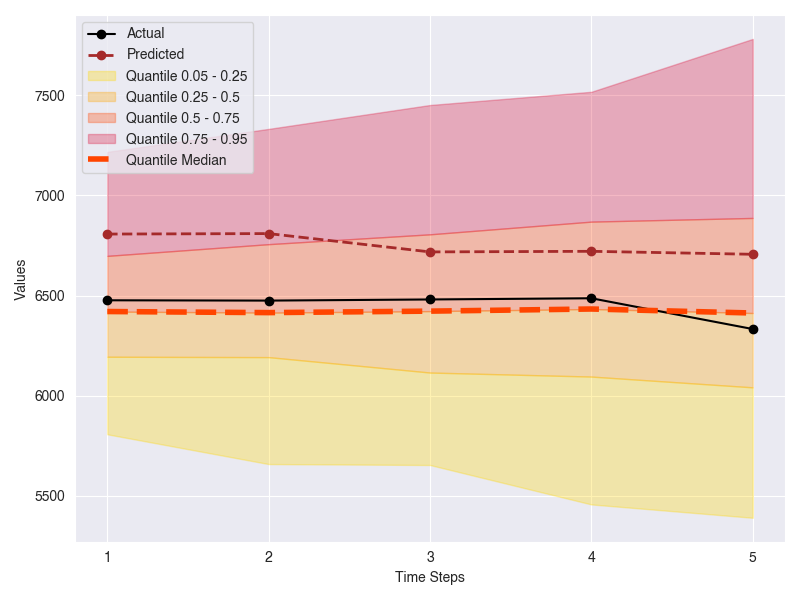}
}
\subfloat[ Multivariate Quantile BD-LSTM]{
    \includegraphics[width=0.49\textwidth]{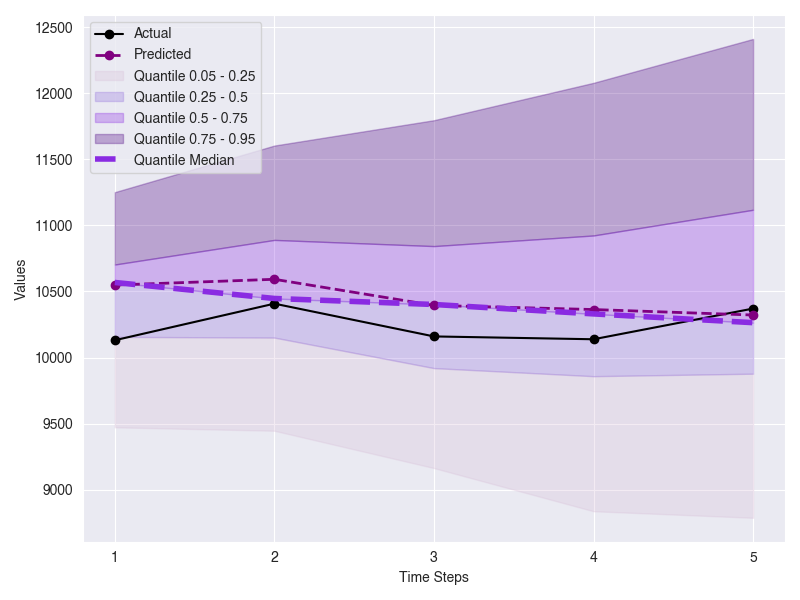}
} \\
\subfloat[Univariate Quantile Conv-LSTM ]{
    \includegraphics[width=0.49\textwidth]{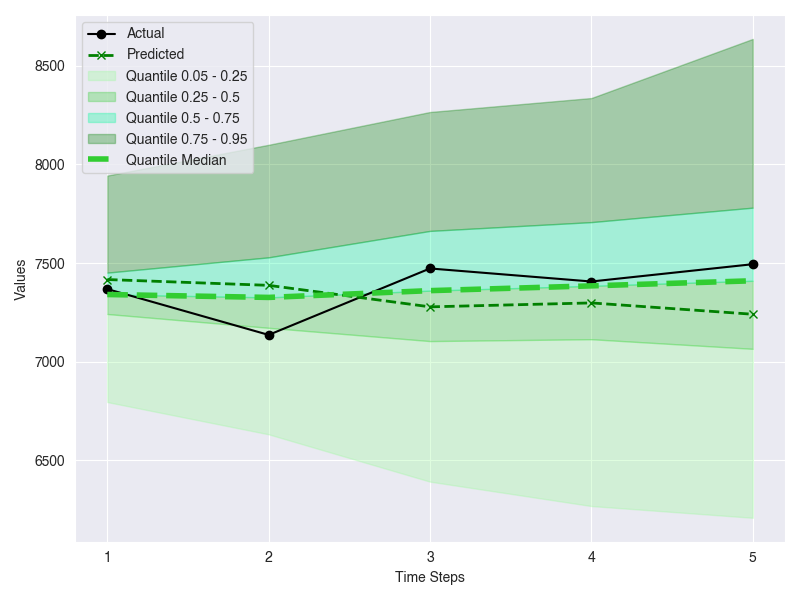}
}
\subfloat[\scriptsize Multivariate Quantile Conv-LSTM]{
    \includegraphics[width=0.49\textwidth]{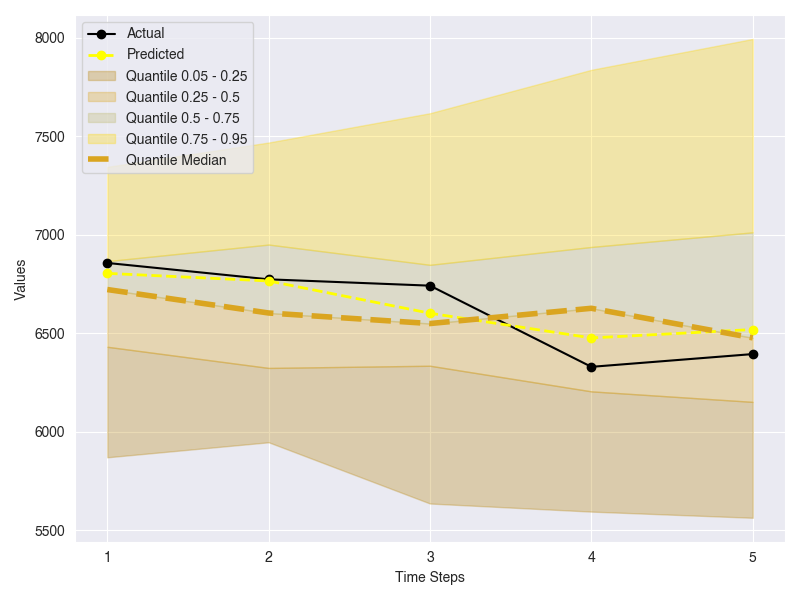}
}\\
\subfloat[Univariate  Quantile ED-LSTM]{
    \includegraphics[width=0.49\textwidth]{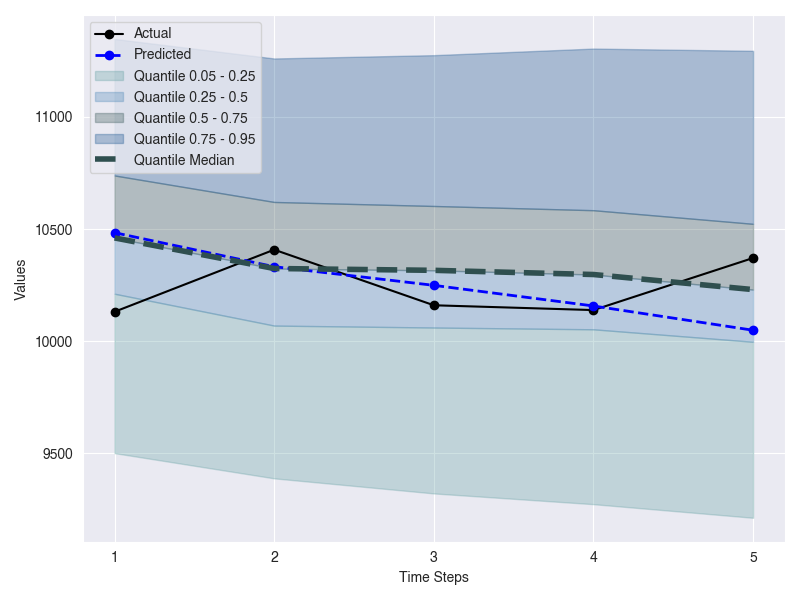}
}
\subfloat[\scriptsize Multivariate Quantile ED-LSTM]{
    \includegraphics[width=0.49\textwidth]{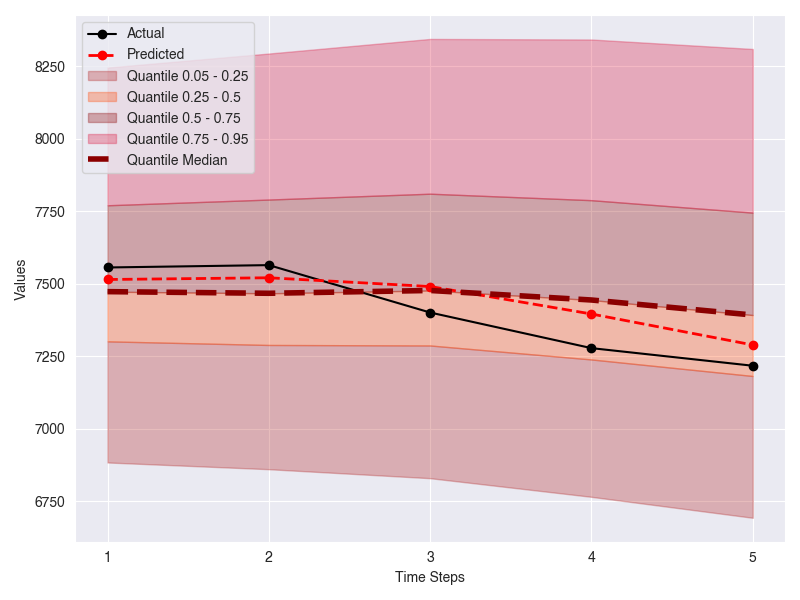}
} \\
\caption{Prediction for the Bitcoin time series, showing quantiles for Univariate and Multivariate strategies for the Quantile-ED-LSTM  (e.g. Quantiles 0.05-0.25) and ED-LSTM (Predicted).}
\label{fig:bitcoipredictions}
\end{figure*}

\begin{figure*}[htbp!]
\subfloat[Univariate Quantile BD-LSTM]{
    \includegraphics[width=0.49\textwidth]{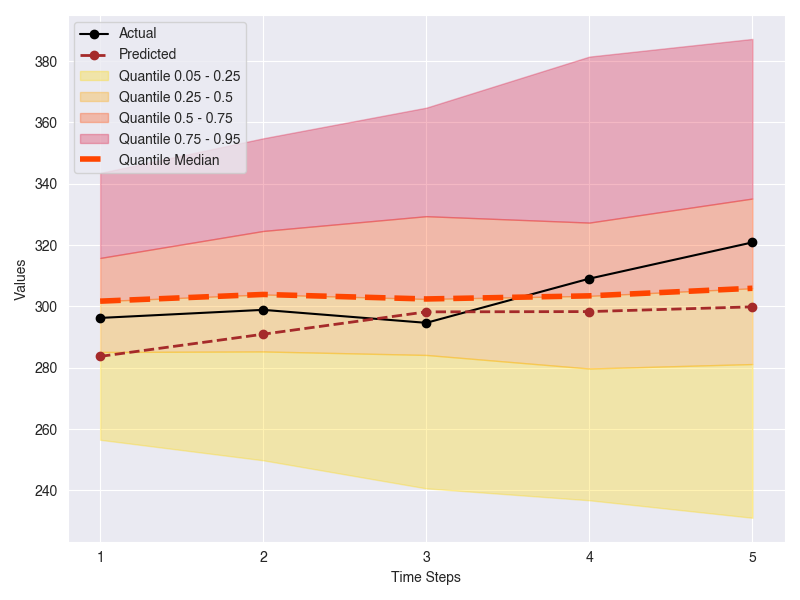}
}
\subfloat[Multivariate Quantile BD-LSTM]{
    \includegraphics[width=0.49\textwidth]{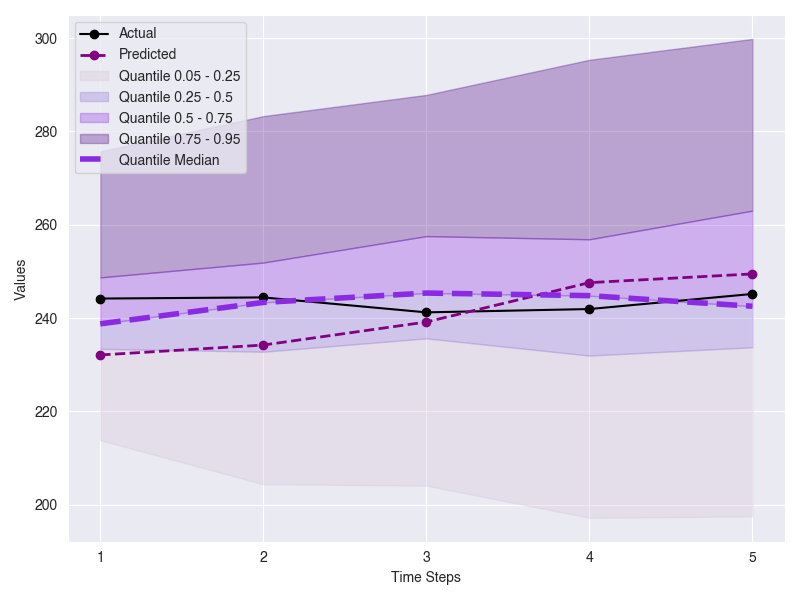}
} \\
\subfloat[Univariate Quantile Conv-LSTM]{
    \includegraphics[width=0.49\textwidth]{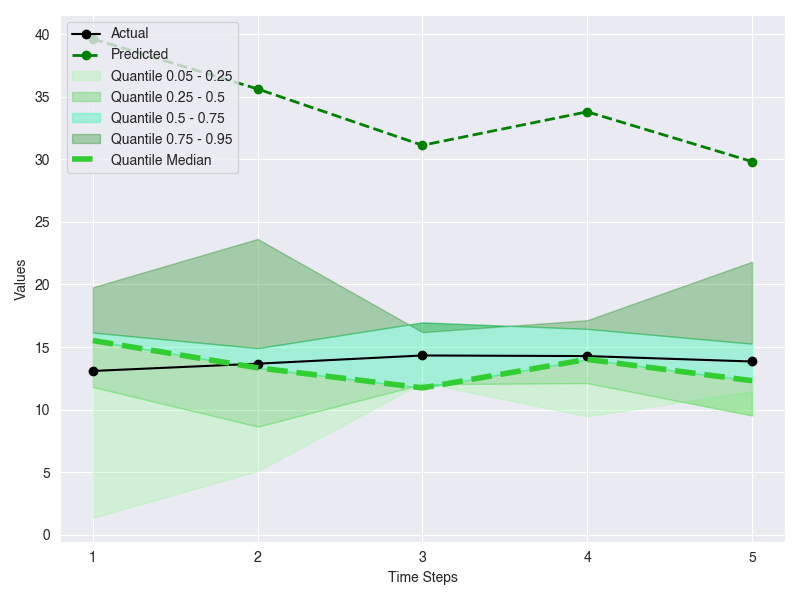}
}
\subfloat[Multivariate Quantile Conv-LSTM]{
    \includegraphics[width=0.49\textwidth]{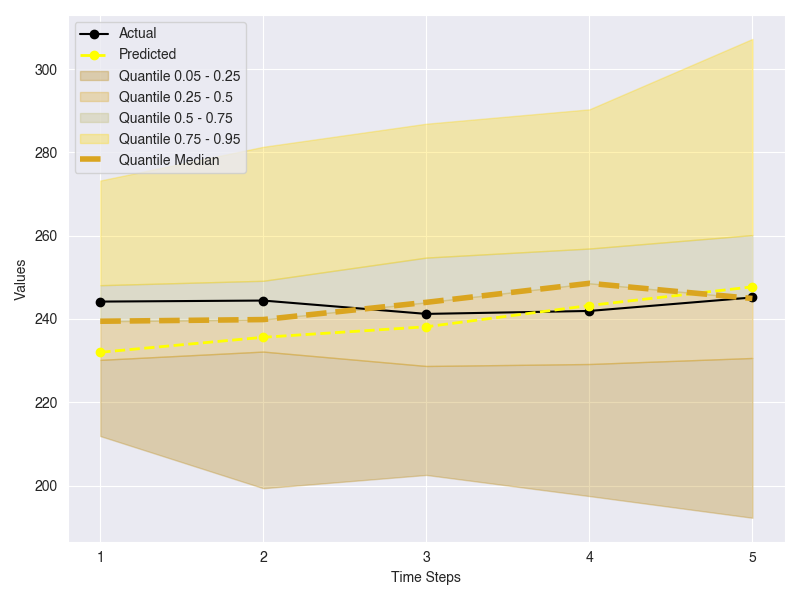}
}\\
\subfloat[Univariate Quantile ED-LSTM]{
    \includegraphics[width=0.49\textwidth]{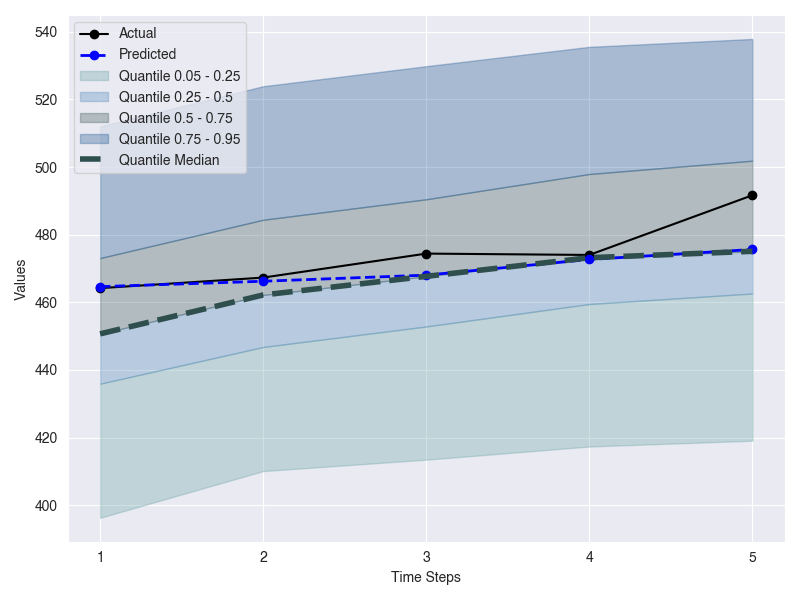}
}
\subfloat[ Multivariate Quantile ED-LSTM]{
    \includegraphics[width=0.49\textwidth]{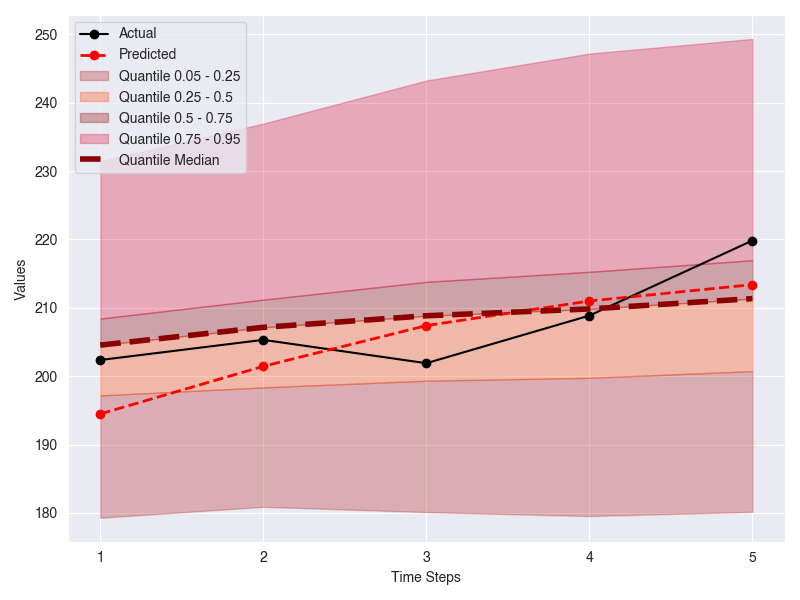}
} \\
 
\caption{Prediction for the Ethereum time series, showing quantiles for Univariate and Multivariate strategies for the Quantile-ED-LSTM (e.g. Quantile 0.05-0.25) and ED-LSTM (Predicted).}
\label{fig:ethpredictions}
\end{figure*}

\begin{figure*}[htbp!]
\subfloat[\scriptsize Sunspot Quantile BD-LSTM]{
    \includegraphics[width=0.49\textwidth]{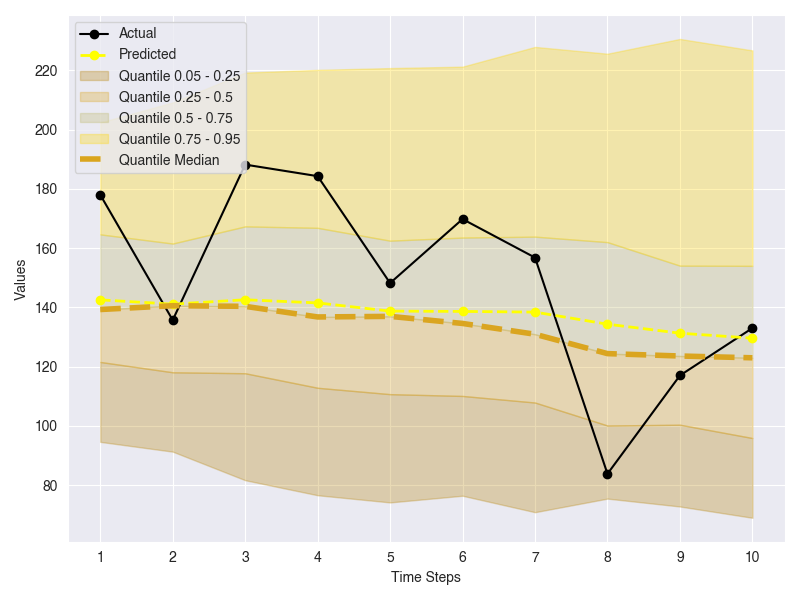}
}
\subfloat[\scriptsize Sunspot Quantile ED-LSTM]{
    \includegraphics[width=0.49\textwidth]{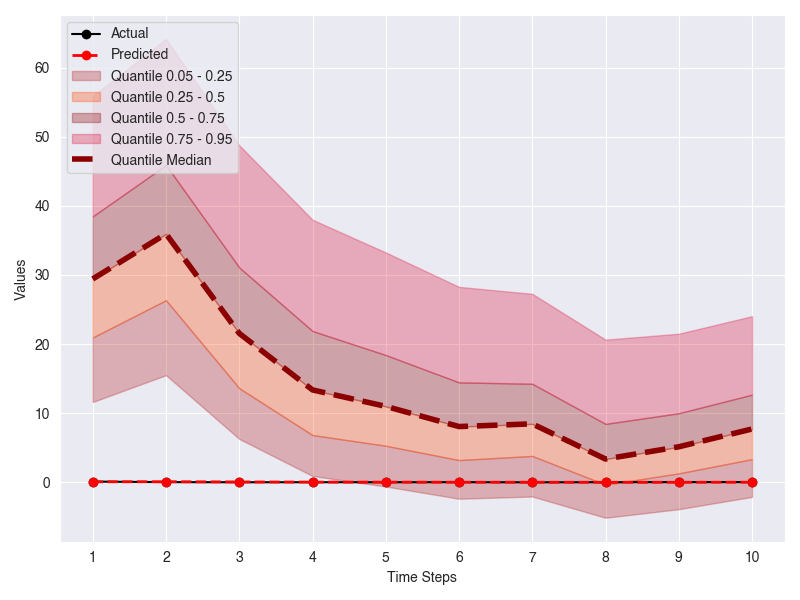}
} \\
\subfloat[\scriptsize Mackey-Glass Quantile BD-LSTM]{
    \includegraphics[width=0.49\textwidth]{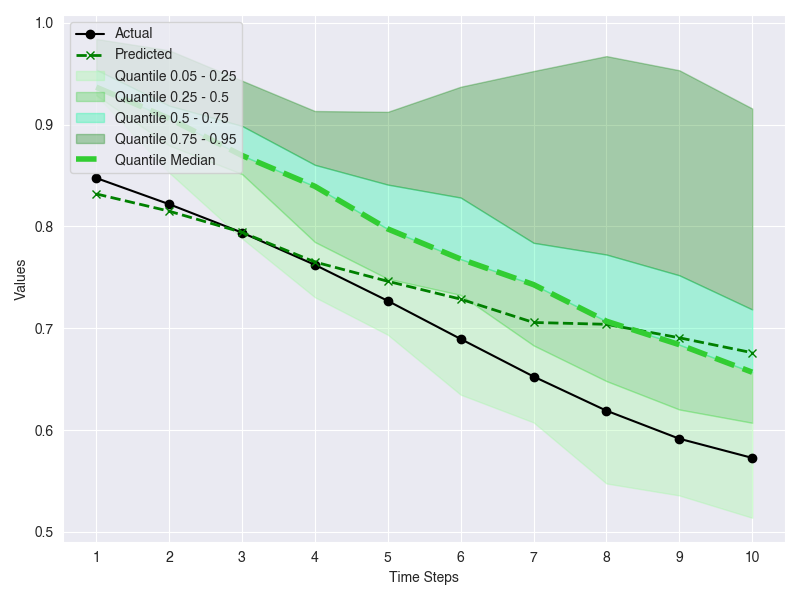}
}
\subfloat[\scriptsize Mackey-Glass Quantile ED-LSTM]{
    \includegraphics[width=0.49\textwidth]{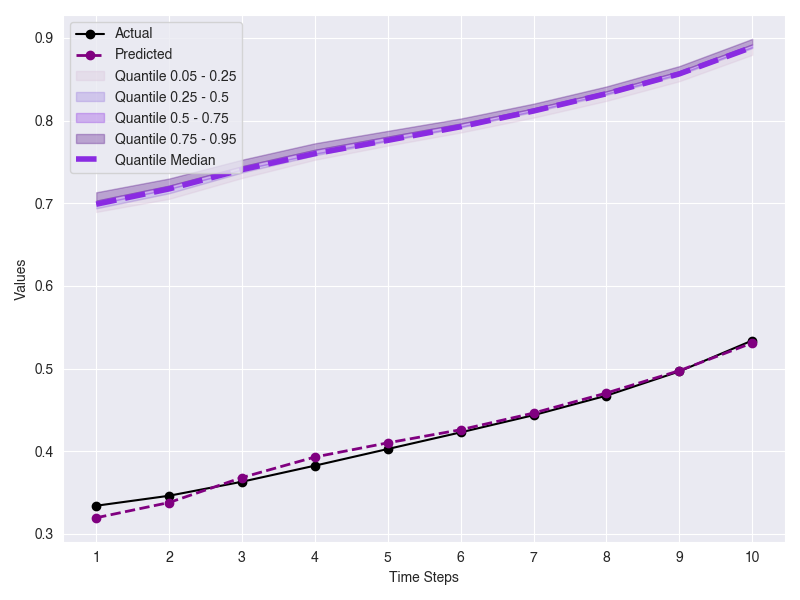}
}\\
\subfloat[\scriptsize Lorenz Quantile BD-LSTM]{
    \includegraphics[width=0.49\textwidth]{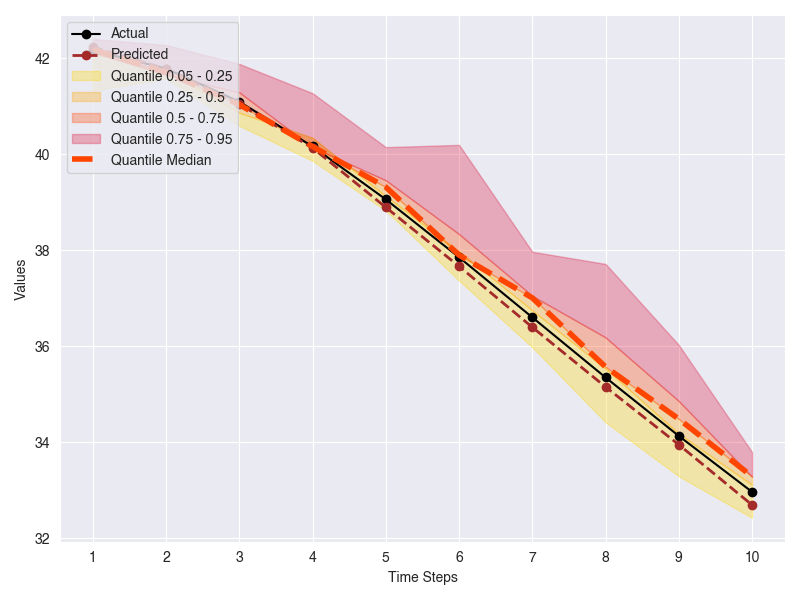}
}
\subfloat[\scriptsize Lorenz Quantile ED-LSTM]{
    \includegraphics[width=0.49\textwidth]{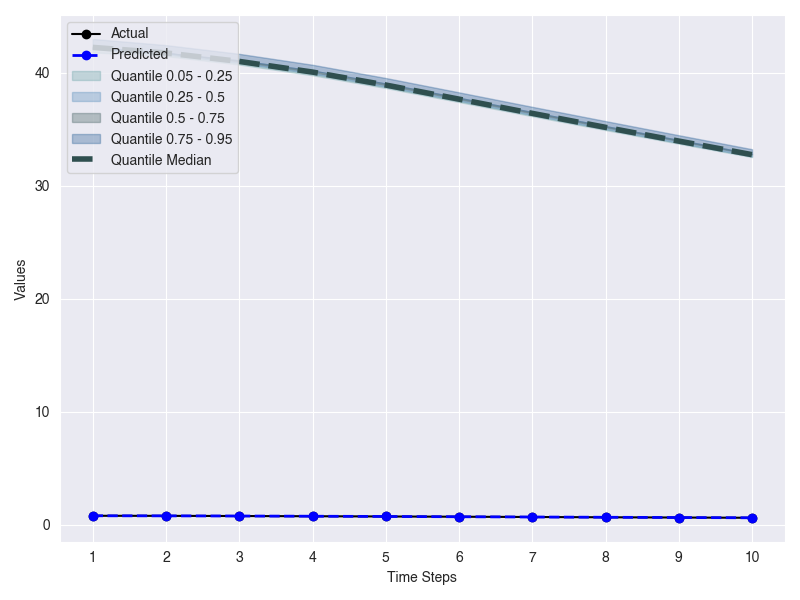}
} \\
\caption{Predictions for univariate Quantile-ED-LSTM models (e.g. Quantile 0.05-0.25) and ED-LSTM (Predicted) for Mackey-Glass, Sunspot and Lorenz time series.}
\label{fig:benchmarkpredictions}
\end{figure*}





\end{document}